\documentclass{elsarticle}

\usepackage[figuresright]{rotating}
\usepackage{graphicx}
\usepackage{subfigure}
\usepackage{multirow}
\usepackage[table,xcdraw]{xcolor}
\usepackage{amsmath}
\usepackage{array}

\journal{J. Comput. Appl. Math.}

\begin{document}
\begin{frontmatter}
	
	\title{Joint Demosaicking and Denoising Benefits from a Two-stage Training Strategy}
	
    \author[address1]{Yu~Guo}\ead{yuguomath@aliyun.com}
    \author[address1]{Qiyu~Jin\corref{cor1}}
    \cortext[cor1]{Corresponding author.}
    \ead{qyjin2015@aliyun.com}
    \author[address2]{Jean-Michel~Morel}\ead{morel@ens-paris-saclay.fr}
    \author[address3]{Tieyong~Zeng}\ead{zeng@math.cuhk.edu.hk}
    \author[address2]{Gabriele~Facciolo}\ead{gabriele.facciolo@ens-paris-saclay.fr}

    \address[address1]{School of Mathematical Science, Inner Mongolia University, Hohhot, China }
    \address[address2]{Centre Borelli, CNRS, ENS Paris-Saclay, Universit\'{e} Paris-Saclay, France.}
    \address[address3]{Department of Mathematics, The Chinese University of Hong Kong, Satin, Hong Kong}

	\begin{abstract}
    Image demosaicking and denoising are the first two key steps of  the color image production pipeline. The classical processing sequence has for a long time consisted of applying denoising first, and then demosaicking. Applying the operations in this order leads to oversmoothing and checkerboard effects. Yet, it was difficult to change this order, because once the image is demosaicked, the statistical properties of the noise are dramatically changed and hard to handle by traditional denoising models. 
    In this paper, we address this problem by a hybrid machine learning method. We invert the traditional color filter array (CFA) processing pipeline by first demosaicking and then denoising.  Our demosaicking algorithm, trained on noiseless images, combines a traditional method and a residual convolutional neural network (CNN). 
    This first stage retains all known information, which is the key point to obtain faithful final results.  The noisy demosaicked  image is then passed through a second CNN restoring a noiseless full-color image. This pipeline order completely avoids checkerboard effects and restores fine image detail.	 Although CNNs can be trained to solve jointly demosaicking-denoising end-to-end, we find that this two-stage training  performs better and is less prone to failure. It is shown experimentally to improve on the state of the art, both quantitatively and in terms of visual quality.  
    \end{abstract}
    
	\begin{keyword}
		Demosaicking, denoising, pipeline, convolutional neural networks, residual.
	\end{keyword}

    \end{frontmatter}

\section{Introduction}

\begin{figure}[t]
	\begin{center}
	\addtolength{\tabcolsep}{-5pt} 
	{%
		\fontsize{8pt}{\baselineskip}\selectfont
		\begin{tabular}{cccc}
		\includegraphics[width=0.23\textwidth]{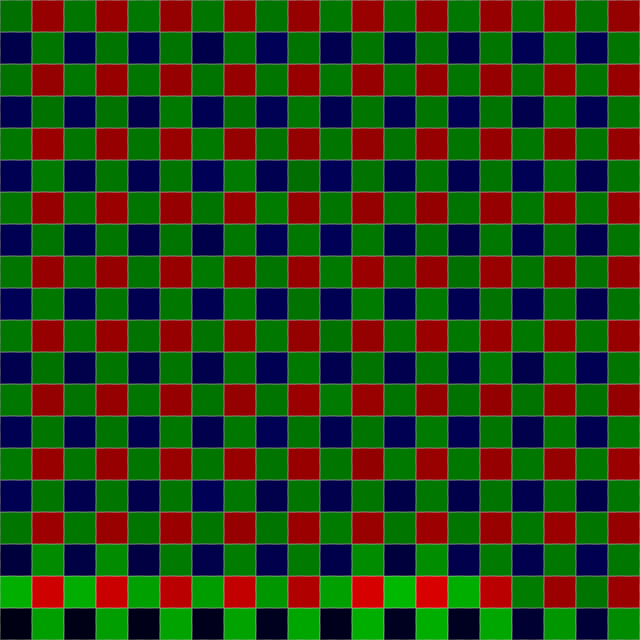} &
		\includegraphics[width=0.23\textwidth]{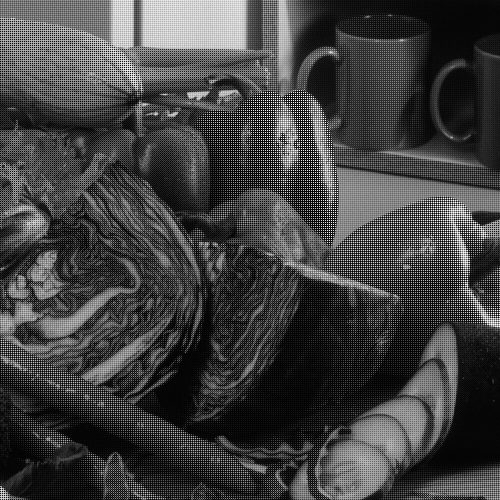} &
		\includegraphics[width=0.23\textwidth]{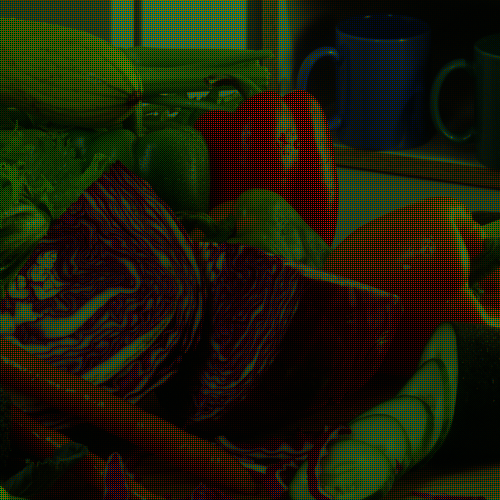} &
		\includegraphics[width=0.23\textwidth]{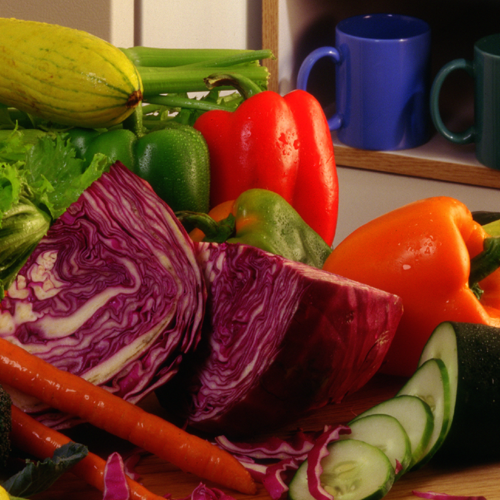} \\
		
		(a) Bayer pattern & (b) CFA image & (c) Mosaic image & (d) Color image \\
		\end{tabular}
		}
	\end{center}
	\caption{The image shows the raw data collected by the sensor and the color filter arrays of the Bayer pattern.}
	\label{Bayer}
\end{figure}

The objective of demosaicking is to build a full-color image from four spatially undersampled color channels. Indeed, digital cameras can only capture one color information on each pixel through a single monochrome sensor, and most of them use color filter arrays (CFA) such as the Bayer pattern \cite{bayer1976color} (shown in Figure \ref{Bayer}) to obtain images. The raw data collected in this way is missing two-thirds of pixels and is contaminated by noise. Hence, image demosaicking, {\em i.e.} the task of reconstructing a full-color image from the incomplete raw data is a typical ill-posed problem.

The conventional method for processing noisy raw sensor data has been to perform denoising and demosaicking as two independent steps. 
Since demosaicking is a complex interpolation process, the raw noise becomes correlated and anisotropic after demosaicking (see~\cite{jin2020review} for a detailed discussion), thus losing its independent Poisson noise structure.
This means that most classic denoising algorithms are not directly applicable. Indeed, most algorithms  rely on the AGWN  (additive Gaussian white noise) assumption, which is approximately valid after a simple Anscombe transform has been applied to the raw data. 
Moreover, most standard demosaicking algorithms with good performance are designed based on the critical noise-free condition.
This takes for granted the assumption that the image processing pipeline starts with denoising~\cite{kalevo2002noise,Park2009beforedemosaicking,lee2017denoising}.

However, some researchers have observed that demosaicking first and then denoising yields a better visual quality. Condat~\cite{Condat2010simple} proposed to demosaick first and then project the noise into the luminance channel of the reconstructed image before denoising according to the grayscale image. This idea was later refined in \cite{Condat2012Joint,Condat2014Generic}. Recently Jin {\em et al.}~\cite{jin2020review}
improved the "demosaicking first" pipeline via a simple modification of the traditional color denoiser, and gave the corresponding theoretical explanation.

Both pipelines have significant shortcomings. A ``denoising first'' pipeline removes noise directly on the CFA image.
Yet, CFA image denoising differs from the usual grayscale or full-color image denoising.  Indeed, CFA image denoising implies subsampling the CFA image into a half-size four-channel RGGB image, which is then  denoised. This leads not only to a poor preservation of image details due to the reduced resolution,  but also to a loss of the correlation between the red (R), green (G) and blue (B) channels. As a result, the restored image is oversmoothed and  checkerboard effects~\cite{Danielyan2009BM3DCFA} are introduced.
On the other hand, the ``demosaicking first'' pipeline also  introduces a thorny issue:
It requires denoising a demosaicked residual noise whose statistical properties have been changed by a complex interpolation, which are hard to model accurately.  
This was almost impossible for traditional denoising algorithms, but current data-driven deep learning based methods offer new paths to solve this problem.
In recent years, deep convolutional neural networks have achieved great success in computer vision and image processing. In image classification and recognition \cite{he2016deep,szegedy2017inception}, denoising \cite{zhang2017Denoiser,Zhang2018FFDNet,Guo2021Fast,Fang2021Multilevel,HOU2022IDPCNN}, demosaicking \cite{gharbi2016deep,tan2017color,Tan2018AdaptiveDemosaicking,Liu_2020_CVPR,Guo2021Joint}, super-resolution \cite{Fang2020Soft-Edge,wen2020residual} and other high-level and low-level visual tasks,  deep learning methods surpass  traditional methods. Since deep learning is data driven, it can find the hidden rules from the data without relying on hand-made filters and a priori knowledge. In this paper, we take advantage of this new flexibility  to handle a noise with complex statistical properties, like the one introduced by a "demosaicking first" pipeline.

We therefore implement a "demosaicking first and  then  denoising" approach  by a  network with a two-stage training strategy. Convolutional neural networks (CNNs) are first combined with traditional algorithms to obtain an effective demosaicking algorithm. Using this demosaicking as a base, we use another CNN to remove the demosaicked residual noise, whose statistical properties have been changed.
Our main contributions are:
\begin{itemize}
	\item 
	 A  CNN architecture implementing the 
	 ``demosaick first and then denoise'' pipeline, which effectively restores full-color images from noisy CFA images while preserving more detail and avoiding oversmoothing and	 checkerboard artifacts. 
	
	\item 
	Ablation studies show that this architecture and the proposed two-stage training strategy perform better than usual end-to-end approaches, enable  a more stable training, and yield state-of-the-art results.
	
   \item
	A modified Inception architecture to implement the two stages of our network.  This choice fosters  cross-channel information fusion for producing a more accurate estimate of the original image and improves the receptive field to reduce artifacts. In that way, we obtain a lighter network than current state-of-the-art approaches~\cite{kokkinos2018deep, xing2021end} without compromising performance.
\end{itemize}

The rest of the paper is organized as follows. 
Section~\ref{sec:relatedwork} presents related work on demosaicking and denoising.
The demosaicking and denoising model is introduced in Section~\ref{sec:Residuallearning}. Section~\ref{sec:Experiments} provides quantitative and qualitative comparisons with state-of-the-art methods. The concluding remarks are given in Section~\ref{sec:Conclusion}.

\section{Related Work}
\label{sec:relatedwork}

\subsection{Demosaicking}
Demosaicking is a classic problem with a vast literature. All authors agree that the key to attaining a good demosaicking is to restore the image regions with high-frequency content. Smooth regions are instead easy to interpolate from the available samples.
The earliest demosaicking algorithms used methods such as spline interpolation and bilinear interpolation to process each channel. These methods introduce serious zipper effects.
In order to eliminate the artifacts at the image edges, Laroche and Prescott~\cite{laroche1994apparatus} introduced a direction adaptive filter by selecting a preferred direction to interpolate the additional color values according to gradient values.
Inspired by this idea, Adams and Hamilton proposed a direction adaptive  inter-channel correlation filter \cite{hamilton1997adaptive,adams1998Proceedings} under the  assumption that derivatives of R, G and B are nearly equal. The G channel interpolation is obtained by a discrete directional Taylor formula involving  the second order derivative   of either the R or the B channel (see \cite{jin2021mathematical}).
Once the G channel interpolation was complete, the G channel was taken as a guide image to help the R and B channel interpolation.
Many advanced algorithms have still extended the idea of a  combination of direction adaptive and  inter-channel correlation. In order to make better use of the correlation between channels, Zhang and Wu~\cite{zhang2005DLMMSE} developed an adaptive filtering method using directional linear minimum mean square error estimation (DLMMSE).
Both horizontal and vertical direction interpolations fail to restore the color value when the pixels are located  near some edge or in textured regions resulting in zipper artifacts at those areas.
In order to solve this issue, Pekkucuksen and Altunbasak~\cite{Pekkucuksen2010gbtf} decomposed the horizontal and vertical directions into four directions of east, west, south, and north on the basis of~\cite{zhang2005DLMMSE}, and then used the color differences in the four directions to estimate the missing G value. 
Similarly to \cite{Pekkucuksen2010gbtf}, Kiku~{\em et al.}~\cite{Kiku2013RI} proposed RI which  calculates four directions'
interpolations of R, G and B channels via a Guided Filter~\cite{He2013Guided}, and improves  the tentative estimates by  substituting a residual technique for the HA interpolation \cite{hamilton1997adaptive,adams1998Proceedings}.
The MLRI~\cite{kiku2014minimized} and MLRI+wei~\cite{Kiku2016WMLRI} were the improved versions of RI  by minimizing the Laplacian energy of the guided filter.
Moreover, ARI~\cite{monno2017adaptive} united the advantages of RI and MLRI by combining both methods  in an iterative process with  the most appropriate number of iteration steps at each pixel.
These last interpolation algorithms  have received a detailed mathematical analysis in \cite{jin2021mathematical}.

In addition to the above local interpolation algorithms,  other classic image processing techniques have been attempted to tackle the problem: algorithms based on non-local similarity~\cite{Buades2009Self-Similarity,Duran2014Self-Similarity,Mairal2009Non-local}, wavelet-based algorithms~\cite{Lu2010Demosaicking,Zhang2018Wavelet}, frequency domain based algorithm~\cite{Dubois2005Frequency-domain,Dubois2006Filter}, and dictionary learning based algorithms~\cite{hua2016context,bai2018demosaicking}.

Accompanying the wide application of deep learning in the field of image processing, demosaicking algorithms based on deep learning achieved great success and redefined the state-of-the-art.
Tan {\em et al.}~\cite{tan2017color}  addressed the demosaicking problem by learning a deep residual CNN.
A two-phase network architecture was designed to reconstruct the G channel first and then estimate the R and B channel using the reconstructed G channel as guide.
After calculating  the inter-channel correlation coefficients, Cui {\em et al.}~\cite{cui20183-part} found that R/G and G/B were more relevant than R/B and established a 3-stage CNN structure  for  demosaicking according to this observation.
Instead of using two-phase or three-phase network architecture,  Tan {\em et  al.}~\cite{Tan2018AdaptiveDemosaicking}  learned directly the residual between ground truth image and an initial full color image obtained by a fast  demosaicking method \cite{malvar2004high}. 
This idea combined the traditional method and CNNs to  simplify the network structure for the demosaicking problem.
Syu {\em et al.}~\cite{syu2018learning} used a convolutional neural network to design a demosaicking algorithm, and compared the effects of convolution kernels of different sizes on the reconstruction. At the same time they also designed a new CFA pattern using a data-driven approach. 
Different from conventional demosaicking CNN methods, Yamaguchi and Ikehara~\cite{Yamaguchi2019Chrominance} took chrominance images as the output of CNN to improve the result.
Higher-Resolution Network (HERN) was proposed by Mei {\em et al.}~\cite{Mei2019HighEr-Resolution} to  solve the demosaicking problem by learning global information from high resolution data with a feasible GPU memory usage.

\subsection{Joint demosaicking and denoising}
Since the raw CFA data is altered by noise while most demosaicking algorithms assume a noise-free image, the image processing pipeline often requires a denoising step. 
Denoising and demosaicking are both ill-posed problems in the pipeline of reconstruction of full color images. In order to reduce artifacts caused by error accumulation, some works have proposed to jointly perform demosaicking and denoising. 
Condat and Mosaddegh~\cite{Condat2012TV} proposed an algorithm based on total variation minimization.
Klatzer {\em et al.}~\cite{Klatzer2016energy}
formulated joint demosaicking and denoising
problem as an energy minimization problem.
Khashabi {\em et al.}~\cite{Khashabi2014Random} introduced a machine learning method by learning a statistical model  from  natural images to avoid artifacts.
Menon and Calvagno~\cite{Menon2009space-varying} evaluated the noise properties  after  the space-varying demosaicking method \cite{meno2009regularization} and then proposed a joint  demosaicking and denoising one.
Tan {\em et al.}~\cite{Tan2017ADMM} addressed the joint  demosaicking and denoising problem as a TV regularization model with multiple effective priors and solved by the alternating direction method of multipliers (ADMM).

The advent of deep learning techniques  and  the increasing availability of large training data sets, have led to a new generation of state-of-the-art algorithms that are able to reconstruct the full color images  from  noisy CFA images.
Gharbi {\em et al.}~\cite{gharbi2016deep}  built a huge image database where images were mined from the web and trained a network on it  for avoiding  zippering or moir\'{e} artifacts.
Inspired by  image regularization methods \cite{lefkimmiatis2018universal}, Kokkinos {\em et al.} \cite{kokkinos2018deep,Kokkinos2019Iterative} established a novel deep learning architecture that combines a 
majorization-minimization algorithm with residual denoising networks. 
Huang {\em et al.} \cite{Huang2018Lightweight} proposed a lightweight end-to-end network using deep residual learning and aggregated residual transformations.
In order to use real data directly, Ehret {\em et al.}~\cite{Ehret2019Joint} introduced a  mosaic-to-mosaic training strategy  which doesn't require ground truth RGB data.  The proposed framework can be used to fine-tune a pretrained network to a RAW burst.
The self-guidance network (SGNet)~\cite{Liu_2020_CVPR} was proposed according to the fact that the G channel of CFA image contains more information. 
The G channel is recovered first and works as a guide image to interpolate the R and B channels.
In~\cite{Guo2021Joint}, G channel prior  features are utilized as  guidance to extract and upsample the features of the whole image.
Xing and Egiazarian~\cite{xing2021end} proposed an end-to-end solution for the joint demosaicking, denoising and super-resolution. They showed that merely training the network with mean absolute error loss function yielded  superior results.

Satisfactory results have been obtained for joint demosaicking and denoising based on deep learning, but these algorithms all rely on the  fitting power of CNNs to solve multiple tasks simultaneously end-to-end. Undoubtedly, this ignores the inter-task correlation, especially the long debated issue of demosaicking and denoising pipeline order.


\section{Residual learning for demosaicking and denoising}
\label{sec:Residuallearning}
The biggest obstacle to applying a demosaicking first and then denoising pipeline is the correlated noise resulting from the demosaicking. This is very difficult for model-based denoisers. Using CNNs can attain satisfactory end-to-end results, however these  methods neglect the dependency between the demosaicking and denoising tasks.
Inspired by \cite{Condat2010simple,jin2020review}, we propose a two-stages CNN for reconstructing full-color images from noisy CFA images.
In the first stage, we design a demosaicking algorithm that combines traditional methods and deep learning by ignoring the noise. All known information is retained in this stage, which is key to obtain good final results.
After the first stage, a noisy full-color image is obtained whose noise statistical properties have been changed by the demosaicking.
The second CNN is used to learn to remove the demosaicked residual noise and to effectively recover the underlying textures.

The noisy CFA model is written as
\begin{equation}\label{Eq noisy cfa}
Y = M.*(X + \varepsilon),
\end{equation}
where $X$ is an original full-color image, $Y$ is the noisy CFA (or mosaicked) image, $\varepsilon$ is Gaussian noise with zero mean and standard deviation $\sigma$, the operator $.*$ denotes the array element-wise multiplication and $M$ denotes the CFA mask. The CFA mask $M$ and its inverse mask are defined as
\begin{equation}
M = \left[
\begin{array}{l}
M_{R} \\
M_{G} \\
M_{B} \\
\end{array}
\right] \quad \text{and} \quad
IM = \left[
\begin{array}{l}
\mathbf{ 1}-M_{R} \\
\mathbf{ 1}-M_{G} \\
\mathbf{ 1}-M_{B} \\
\end{array}
\right],
\end{equation}
\begin{equation*}
M_{R}(i,j)=\left\{
\begin{array}{ll}
0, &\text{if } (i,j)\notin \Omega_{R}; \\
1, &\text{if } (i,j)\in \Omega_{R},
\end{array}
\right.
\end{equation*}
\begin{equation*}
M_{G}(i,j)=\left\{
\begin{array}{ll}
0, &\text{if } (i,j)\notin \Omega_{G}; \\
1, &\text{if } (i,j)\in \Omega_{G},
\end{array}
\right.
\end{equation*}
\begin{equation*}
M_{B}(i,j)=\left\{
\begin{array}{ll}
0, &\text{if } (i,j)\notin \Omega_{B}; \\
1, &\text{if } (i,j)\in \Omega_{B},
\end{array}
\right.
\end{equation*}
where $\mathbf{ 1}(i,j)=1$, $\Omega$ denotes the set of CFA image pixels, $\Omega_{R}$, $\Omega_{G}$, $\Omega_{B}\subseteq\Omega$ are disjoint sets of pixels, which respectively record R, G and B values in the CFA image, and satisfy $\Omega_{R}\cup\Omega_{G}\cup\Omega_{B}=\Omega$.

\begin{figure*}[ht]
	\centering
	\includegraphics[width=0.9\textwidth]{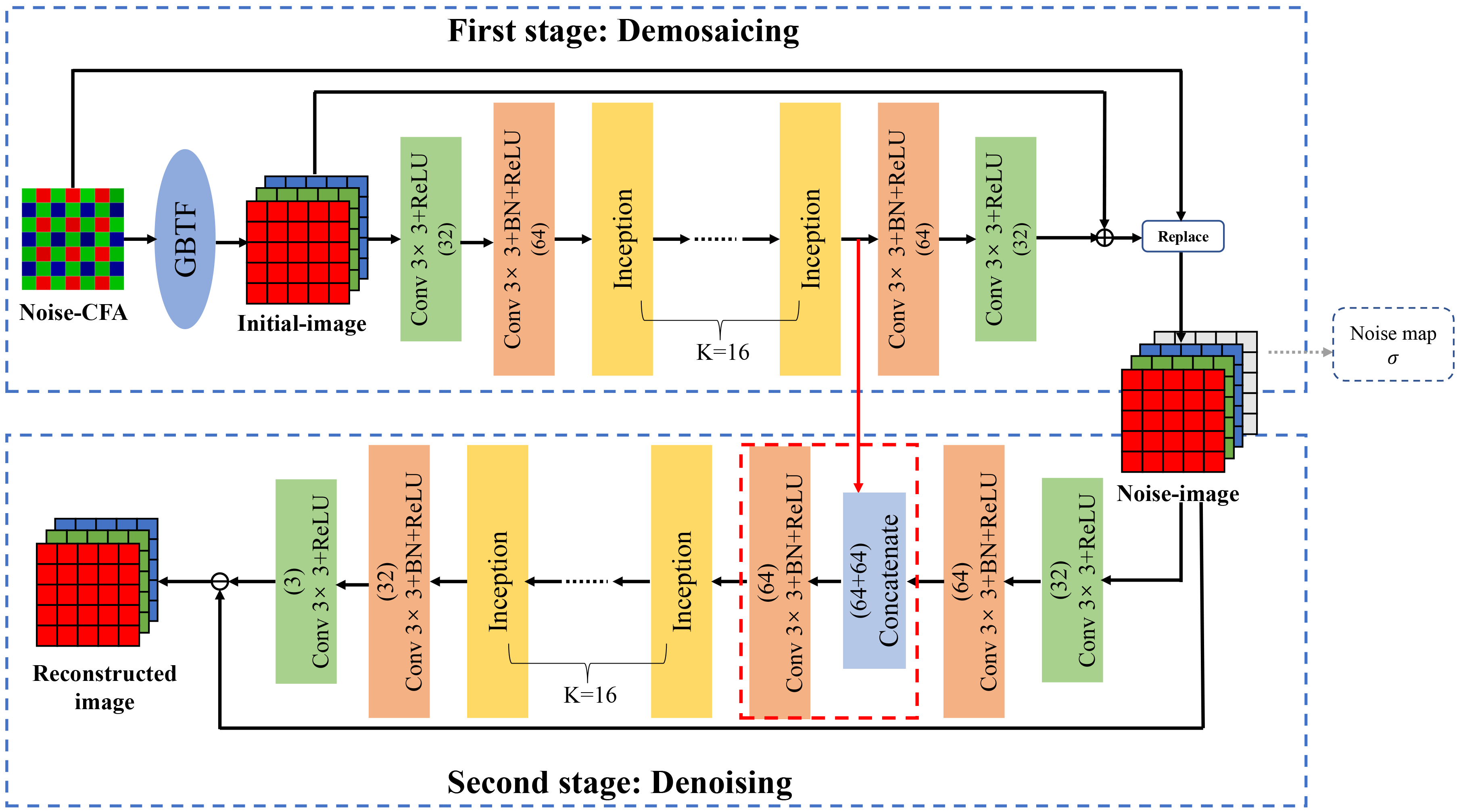}
	\caption{Our two stages CNN architecture for demosaicking-denoising. The first stage takes GBTF to preprocess the CFA image and uses a CNN to learn   residuals  improving the demosaicking  performance of  GBTF. In the second stage, when the noisy CFA image is demosaicked, another CNN is used to learn the residual noise in order to reconstruct the finally full-color image. The term "replace" corresponds to Eq. (\ref{eq: model demosaicking}).}
	\label{Algorithm}
\end{figure*}

The first stage considers only the noise-free CFA model
\begin{equation}
Y = M.*X,
\end{equation}
where $X$ is a full-color image, $Y$ is the noise-free CFA (or mosaicked) image.
We first use the GBTF algorithm~\cite{Pekkucuksen2010gbtf} to obtain a raw demosaicked image
$\widehat{X}_{GBTF} = \mathrm{GBTF}(Y)$ and a residual $R_{GBTF} = X - \widehat{X}_{GBTF}$.
The residual is then corrected with a CNN. For that we use a modified Inception architecture in order to achieve better performance in learning the residual and get an estimator $\widehat{R}_{GBTF}$ (see Figure \ref{Algorithm}).

The final full-color image is obtained as
\begin{equation} \label{eq: model demosaicking}
\widehat{X}_{DM} 
= IM.*(\widehat{X}_{GBTF} + \widehat{R}_{GBTF}) + M.* X.
\end{equation}
The first term in the above equation is the demosaicked image estimated by the CNN and evaluated on the inverse CFA mask $IM$, while the second term is unaltered input CFA samples on the mask $M$.
The resulting CNN is adapted to demosaick noise-free images. So, applying it to a noisy CFA image, produces a noisy demosaicked image.

To handle noisy CFA images, another stage is needed to remove the noise.
Given the trained demosaicking as a basic component, we apply it to model~\eqref{Eq noisy cfa} and obtain a noisy full-color image $\widehat{X}_{DM}$ which
can be decomposed as
\begin{equation} \label{Eq: model noisy fc image}
\widehat{X}_{DM} = X + \varepsilon_{DM}.
\end{equation}
Here, $\varepsilon_{DM}$ is the residual noise (including artifacts) of the demosaicked image, which is no longer independent identically distributed (I.I.D.), 
and has complex unknown statistical properties.
This would be extremely challenging for 
traditional denoising models that strongly rely on statistical assumptions, 
{therefore} we use another CNN to learn 
{to extract the} residual noise $\varepsilon_{DM}$ and obtain the estimator $\widehat{\varepsilon}_{DM}$ (see Figure~\ref{Algorithm}).
The final full-color image is reconstructed as 
\begin{equation}
\widehat{X}_{DMDN} = \widehat{X}_{DM} - \widehat{\varepsilon}_{DM}.
\end{equation}

There are several advantages in training separate demosaicking and denoising networks:
\begin{itemize}
	\item First, the noise-free demosaicking focuses on reconstructing the structure and details in the image without concessions. In addition, the demosaicking network needs not be adapted to each noise level, and all known information is preserved. 
    \item Second, the demosaicked result facilitates the task of the denoiser which has to adapt only to the noise and demosaicking artifacts. As we will see later, training a joint denoising and demosaicking network with equivalent capacity as the demosaicking and denoising networks indeed yields lower quality results. 
    \item Third, the proposed two stage architecture and trainig strategy is more stable at training time than an end-to-end network with equal capacity.
\end{itemize}

\subsection{Demosaicking in a noise-free setting}
The CFA images are different from ordinary images as the values of adjacent pixels represent the intensity of different colors. Many of the existing deep learning algorithms subsample the CFA images to four-channel RGGB images and send them to the network. However, this operation reduces the image resolution. Therefore, the network needs to perform functions similar to super-resolution, and cannot only focus on image demosaicking.
Some algorithms use bilinear interpolation as preprocessing in order to preserve the spatial arrangement of the samples.
However, the bilinear interpolation results are  suboptimal and this affects the performance of the convolutional network.
In this work, we use the gradient based threshold free (GBTF) method \cite{Pekkucuksen2010gbtf}, which has superior performance compared to the bilinear interpolation. Improving the network input also alleviates the task for the network. In subsequent ablation experiments,  GBTF is shown to better preserve image textures.

\begin{figure}[t]
	\centering
	\includegraphics[width=0.7\textwidth]{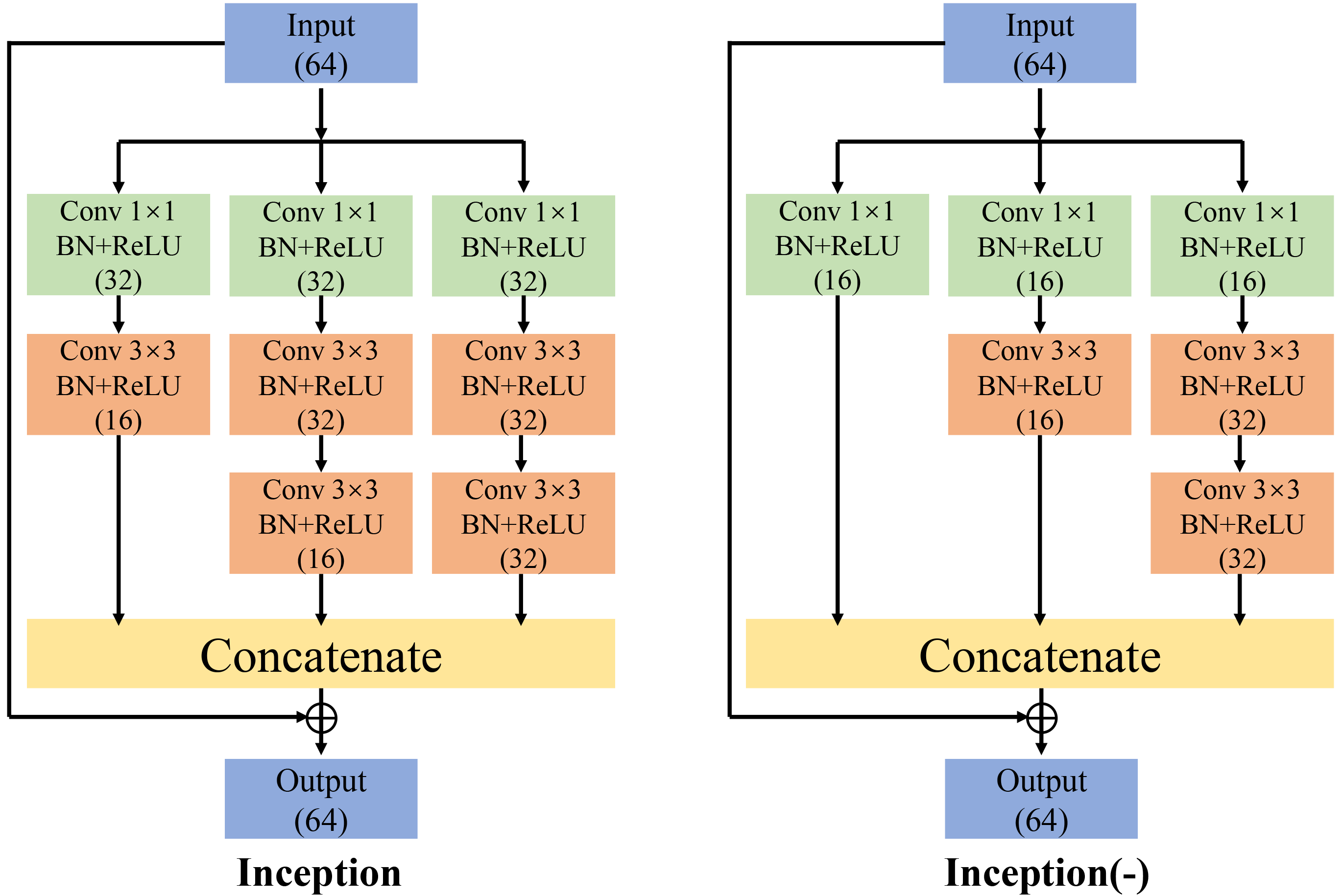} \\
	\caption{Architecture of the Inception block. In order to get a better cross-channel fusion and a larger receptive field, we use $1\times1$ convolution kernels and three-way branches to reduce the parameters while strengthening the fusion of cross-channel information. This is extremely important for demosaicking.}
	\label{Inception}
\end{figure}

After the CFA image is preprocessed, we use a convolutional neural network for residual learning. 
The network architecture is shown in Figure~\ref{Algorithm}.
Syu {\em et al.} pointed out in their work~\cite{syu2018learning} that convolution kernels of different sizes will affect the reconstruction accuracy. The larger the size of the convolution kernel, the higher the reconstruction accuracy. However, the number of parameters using a $5\times5$ convolution kernel is 2.7 times that of using a $3\times3$ convolution kernel. 
We want to increase the receptive field, but without giving up the lightweight $3\times3$ convolution kernels.
In the image demosaicking task, due to the lack of color information, the full-color image reconstruction must make full use of the correlation of the three RGB channels. Therefore, the degree of cross-channel information fusion determines the performance of the demosaicking algorithm.
In order to get a better cross-channel fusion and a larger receptive field, we propose  modifying the architecture of GoogleNet Inception-ResNet~\cite{szegedy2017inception} and adapting the Inception block.  
On the one hand, it is scalable and can increase the receptive field of the network without increasing the number of parameters and computations. On the other hand, the multi-branch structure facilitates the extraction and fusion of features at different levels.
The proposed network has 16 Inception blocks.
The architecture of the Inception block and a lightweight version we propose in this paper are shown in Figure~\ref{Inception}.
In the Inception block, we use $1 \times 1$ convolution kernels to fuse and compress the channels, and use three-way branches to learn different residual features, and finally concatenate the three-way branches.
We also design a lightweight Inception block, which will be denoted by (-) in what follows. 
With roughly the same number of parameters as a $3 \times 3$ Conv-BN-ReLU block for 64-layer feature maps, the proposed Inception block increases the network depth (3 non-linearities) and has a larger receptive field ($5 \times 5$). Moreover, the Inception(-) uses about 50$\%$ of the parameters of the $3 \times 3$ Conv-BN-ReLU. The parameter comparison data are shown in Table~\ref{Inception Compare}. 

\begin{table}[t]
	\begin{center}
	\caption{Inception architecture and number of parameters. The depth of Conv-BN-ReLU is 1, the receptive field is 3, but the depth of the Inception is 3, and receptive field is 5. And Inception(-) has the same properties and the number of parameters is only $52.8\%$ of Conv-BN-ReLU.}
	\label{Inception Compare}
	\scriptsize
	\renewcommand\arraystretch{1.2}
		\begin{tabular}{|m{1.9cm}<{\centering}|m{1.3cm}<{\centering}|m{1.7cm}<{\centering}|m{1.6cm}<{\centering}|}
			\hline
			& Inception & Inception(-) &  Conv.    \\ \hline
			Input the number of feature layers  & 64 & 64 & 64    \\ \hline
			\multirow{2}{*}{First branch}  & 32($1\times1$) & \multirow{2}{*}{16($1\times1$)}  & \multirow{8}{*}{$3\times3$} \\
			& 16($1\times1$) &  &  \\ \cline{1-3}
			\multirow{3}{*}{Second branch}   & 32($1\times1$) & \multirow{3}{*}{\begin{tabular}[c]{@{}c@{}}16($1\times1$) \\ 16($3\times3$) \end{tabular}} &  \\
			& 32($3\times3$)  &  &   \\
			& 16($3\times3$)  &  &   \\ \cline{1-3}
			\multirow{3}{*}{Third branch}   & 32($1\times1$) & 16($1\times1$) &  \\
			& 32($3\times3$)   & 32($3\times3$)   &   \\
			& 32($3\times3$)   & 32($3\times3$)   &   \\ \hline
			Output the number of feature layers & 64 & 64  & 64    \\ \hline
			Number of parameters  & 39360 & 19456  & 36992  \\ \hline
			GFLOPs {} ($128\times128$)  & 0.649 & 0.321 & 0.607  \\ \hline
		\end{tabular}
	\end{center}	
\end{table}

\subsection{Denoising after demosaicking}

Since the  demosaicking stage is trained in a noise-free setting, when a noisy input is demosaicked its output will contain a correlated residual noise.
Removing this noise  also requires  learning. We therefore propose to use a denoising network to remove the structured noise resulting from demosaicking a noisy CFA image. We first   learn a network for each noise level, but in the experiment sections we will also consider a noise level flexible network trained on a range of noise levels (with $\sigma \in [0,20]$ as in~\cite{gharbi2016deep}). In the noise level flexible network, the noise map shown in Figure~\ref{Algorithm} is introduced, which consists of the standard deviation $\sigma$ of Gaussian noise added to the CFA image.

For the denoising network, we also use the same Inception block architecture as the demosaicking network. As shown in Figure~\ref{Algorithm}, the demosaicked image is used as input to the denoising stage. In addition, the features computed at the last layer of the demosaicking stage are reused by introducing them into the denoising stage by a skip connection.

\subsection{Training procedure}
The two stages of our method are trained independently, each with its own loss, which are both based on the classical mean square error (MSE) loss. In the first stage, the network is trained on a noise-free dataset. The loss for the noise-free demosaicking stage is
\begin{equation}
\begin{aligned}
\mathcal{L}_{DM}(\Theta_{DM})=\frac{1}{2N}\sum_{i=1}^N\begin{Vmatrix} \widehat{X}_{DM}^{i}-X^{i} \end{Vmatrix}^2,
\end{aligned}
\end{equation}
\begin{equation}
\widehat{X}_{DM}^{i} = IM .* \left(\widehat{X}_{GBTF}^{i} + F(\widehat{X}_{GBTF}^{i};\Theta_{DM})\right) + M .* \widehat{X}^{i},
\end{equation}
where $F(\widehat{X}_{GBTF}^{i};\Theta_{DM})$ is the output of the demosaicking network to estimate the residual ${R}_{GBTF}$ (see \eqref{eq: model demosaicking}).

After the demosaicking network is trained, we apply it to noisy CFA images (see model \eqref{Eq noisy cfa}) to produce  noisy full-color images (see model \eqref{Eq: model noisy fc image}). The goal of 
the second stage is then to remove residual demosaicked noise  $\varepsilon_{DM}$. 
Therefore the loss for this stage is
\begin{equation}
\mathcal{L}_{DN}(\Theta_{DN})=\frac{1}{2N}\sum_{i=1}^N\begin{Vmatrix} \widehat{X}_{DMDN}^{i} -X^{i} \end{Vmatrix}^2,
\end{equation}
\begin{equation}
\widehat{X}_{DMDN}^{i} = \widehat{X}_{DM}^{i} - G(\widehat{X}_{DM}^{i};\Theta_{DN}),
\end{equation}
where $G(\widehat{X}_{DM}^{i};\Theta_{DN})$ is the output of the denoising network, which works as an estimator of $\varepsilon_{DM}$.

For training the joint demosaicking and denoising, Gharbi {\em et al.} provided a dataset of two million $128\times128$ images (MIT Dataset)~\cite{gharbi2016deep}.
Ma {\em et al.} established the Waterloo Exploration Database (WED) with 4,744 high-quality natural images~\cite{Ma2017Waterloo} and Syu {\em et al.} provided the Flickr500 with 500 high-quality images~\cite{syu2018learning}.
We use these datasets to build our training and test sets.
Indeed, 100,000 images were randomly selected from the MIT dataset. And 4653 images in WED and 491 images in Flickr500 were randomly cropped into 100,000 images ($128\times128$).
These 200,000 patches ($128\times128$) constitute our training set.
Furthermore, 91 images in WED and 9 images in Flickr500 composed our test set.
During the training time, the patch was flipped and rotated $180^{\circ}$ with a 50$\%$ probability for data augmentation.

For training the denoising model we started by adding Gaussian white noise to the CFA images sampled from the training set (see Table~\ref{DN_Kodak} for the standard deviation $\sigma$ of the noise) and applied the demosaicking network to the noisy CFA images. 
The color residual noise images, which were obtained by feeding the noisy CFA images into the demosaicing network, were utilized for training the denoising model.

The network architecture was implemented in PyTorch. 
The network weights were initialized using~\cite{He2015Delving} and the biases were first set to 0.
The optimization was performed by the ADAM optimizer~\cite{kingma2014adam} using the default parameters.
The batch size was set to 64, and the initial learning rate to $10^{-2}$. 
The learning rate decay strategy was the exponential decay method, and the learning rate decayed by 0.9 every 3000 iterations.
Our model was trained on a NVIDIA Tesla V100 and required 50 epochs for each training iteration. The non-lightweight demosaicing and denoising algorithms at each level typically took approximately 3 days to train, while the lightweight algorithms could be trained within a day.

\section{EXPERIMENTS}
\label{sec:Experiments}

\subsection{Datasets}
We chose the classic Kodak \cite{zhang2011color} and McMaster \cite{Dubois2005Frequency-domain} datasets for evaluating our algorithm on the demosaicking and denoising task.
The Kodak dataset consists of 24 images ($768\times512$).
The McMaster dataset consists of 18 images ($500\times500$), which were cropped from the $2310\times1814$ high-resolution images.
At the same time, we conducted experiments on our test set, Urban100 dataset \cite{Huang2015Single} and MIT moiré \cite{gharbi2016deep} to verify the reliability of our proposed algorithm. The Urban100 dataset is often used in super-resolution tasks and contains 100 high-resolution images.
MIT moiré is the test set used by the JCNN algorithm \cite{gharbi2016deep}, which contains 1000 images of $128\times128$ resolution that are prone to generate moiré.

\subsection{Quantitative and qualitative comparisons}
Peak signal-to-noise ratio (PSNR)~\cite{Alleysson2005CPSNR} and structural similarity (SSIM)~\cite{Wang2004SSIM} were used to evaluate the performance of the algorithms.

\begin{sidewaystable}[htpb]
	\centering
	\caption{Comparison with state-of-the-art algorithms in noise-free demosaicking. The best value is marked in {\textbf{bold}}, the second is marked in {\color{red}red}, and the third is marked in {\color{blue}blue}. In the table, (-) indicates a lightweight version.}
	\label{DM}
	\footnotesize
	\renewcommand\arraystretch{1.2}
	\begin{tabular}{|l|c|c|c|c|c|c|}
		\hline
		& Kodak  & McMaster & WED+Flickr & Urban100 & MIT moiré & Average \\ \cline{2-7} 
		\multirow{-2}{*}{Algorithm}       & PSNR/SSIM  & PSNR/SSIM  & PSNR/SSIM & PSNR/SSIM & PSNR/SSIM & PSNR/SSIM\\ \hline
		GBTF \cite{Pekkucuksen2010gbtf} & 40.62/0.9859 & 34.38/0.9322 & 36.35/0.9664 & 34.82/0.9701 & 32.18/0.9120 & 35.67/0.9533 \\ \hline
		MLRI+wei \cite{Kiku2016WMLRI} & 40.26/0.9850 & 36.89/0.9620 & 36.76/0.9707 & 34.90/0.9732 & 32.17/0.9119 & 36.20/0.9606 \\ \hline
		ARI \cite{monno2017adaptive} & 39.91/0.9815 & 37.57/0.9654 & 37.46/0.9745  & 35.35/0.9751 & 32.60/0.9162 & 36.58/0.9625 \\ \hline
		C-RCNN \cite{kokkinos2018deep} & 39.93/0.9843 & 36.68/0.9509 & 37.57/0.9725 & 37.16/0.9797 & 32.99/0.9148 & 36.87/0.9604 \\ \hline
		JCNN \cite{gharbi2016deep} & 42.09/0.9881 & 38.95/0.9695 & 39.24/0.9807 & 38.12/0.9842 & {\textbf{36.65}}/{\textbf{0.9588}} & 39.01/0.9763 \\ \hline
		CDM-CNN \cite{tan2017color} & 41.98/0.9879 & 38.94/0.9696 & 39.52/0.9812 & 38.09/0.9836 & 34.28/0.9311 & 38.56/0.9707 \\ \hline
		CDM-3-Stage \cite{cui20183-part} & 42.31/0.9885 & {\color{red}39.34}/{\color{red}0.9716} & {\color{red}40.12}/{\color{red}0.9827} & {\color{blue}38.60}/{\color{blue}0.9849} & 34.78/0.9334 & {\color{blue}39.03}/{\color{blue}0.9722} \\ \hline
		LCNN-DD \cite{Huang2018Lightweight} & {\color{blue}42.42}/0.9886 & 39.07/0.9701 & 39.75/0.9817 & 38.37/0.9841 & 34.78/0.9338 & 38.88/0.9717 \\ \hline
		JDNDMSR \cite{xing2021end} & 42.35/{\color{red}0.9891} & 38.83/0.9680 & 39.44/0.9810 & 38.33/0.9839 & 35.36/0.9338 & 38.86/0.9712 \\ \hline
		Ours(-) & {\color{red}42.49}/{\color{blue}0.9888} & {\color{blue}39.25}/{\color{blue}0.9702} & {\color{blue}39.84}/{\color{blue}0.9820} & {\color{red}38.88}/{\color{red}0.9852} & {\color{blue}35.97}/{\color{blue}0.9516} & {\color{red}39.29}/{\color{red}0.9756} \\ \hline
		Ours    & {\textbf{42.76}}/{\textbf{0.9893}} & {\textbf{39.61}}/{\textbf{0.9725}} & {\textbf{40.22}}/{\textbf{0.9831}} & {\textbf{39.52}}/{\textbf{0.9864}} & {\color{red}36.53}/{\color{red}0.9533} & {\textbf{39.73}}/{\textbf{0.9769}} \\ \hline
	\end{tabular}
\end{sidewaystable}

\paragraph{Noise-free demosaicking}
In the noise-free CFA image demosaicking task, we compared three traditional algorithms (GBTF~\cite{Pekkucuksen2010gbtf}, MLRI+wei~\cite{Kiku2016WMLRI}, ARI~\cite{monno2017adaptive}) and six deep learning algorithms (C-RCNN~\cite{kokkinos2018deep}, JCNN~\cite{gharbi2016deep}, CDM-CNN~\cite{tan2017color}, CDM-3-Stage~\cite{cui20183-part}, LCNN-DD~\cite{Huang2018Lightweight}, JDNDMSR~\cite{xing2021end}). 
Table~\ref{DM} summarizes the performance of all algorithms on the dataset.
We can see that our proposed algorithm outperforms the other algorithms in the noise-free demosaicking.
On the Kodak dataset, our proposed method surpasses the state-of-the-art by 0.34 dB in the average PSNR value. This gain is 0.27 dB on the McMaster dataset and 0.92 dB on Urban100. At the same time, our proposed lightweight method also achieved good performance. It ranks second on the Kodak and Urban100 dataset and third on the McMaster dataset. On the MIT moiré, the average PSNR value of our proposed algorithm is 0.12 dB lower than that of JCNN~\cite{gharbi2016deep}, but we only used $5\%$ of the training data they provided.

\begin{figure*}[t]
	\begin{center}
	    \tiny
		\renewcommand{\arraystretch}{0.5} \addtolength{\tabcolsep}{-5pt} {%
			\begin{tabular}{cccccc}
				\multirow{10}{*}[10pt]{\includegraphics[width=0.25\textwidth]{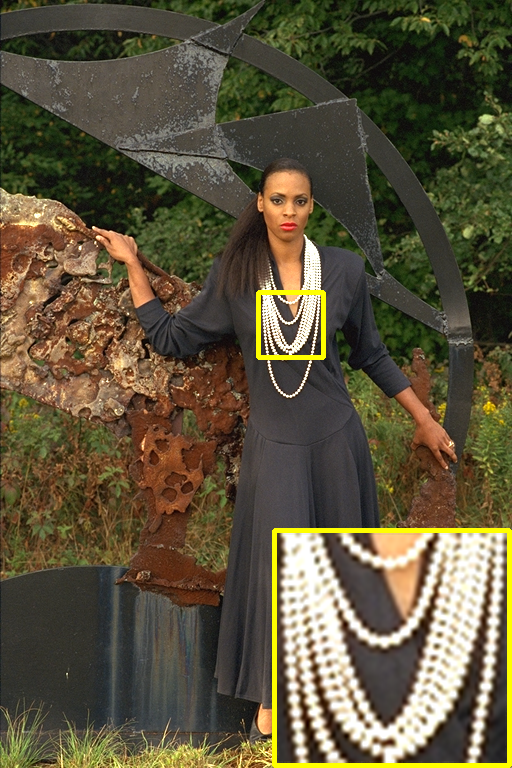}} & 
				\includegraphics[width=0.14\textwidth]{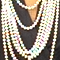} & 
				\includegraphics[width=0.14\textwidth]{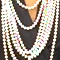} & 
				\includegraphics[width=0.14\textwidth]{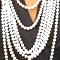} & 
				\includegraphics[width=0.14\textwidth]{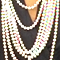} & 
				\includegraphics[width=0.14\textwidth]{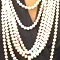} \\

				{} & 
				\includegraphics[width=0.14\textwidth]{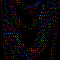} & 
				\includegraphics[width=0.14\textwidth]{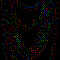} & 
				\includegraphics[width=0.14\textwidth]{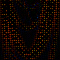} & 
				\includegraphics[width=0.14\textwidth]{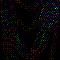} & 
				\includegraphics[width=0.14\textwidth]{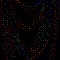} \\
				
				{} & GBTF & ARI& C-RCNN & JCNN & CDM-CNN \\
				{} & 37.85/0.9825 & 37.07/0.9754 & 36.40/0.9776 & 38.42/0.9835 & 38.92/0.9839 \\
				
				{} & 
				\includegraphics[width=0.14\textwidth]{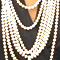} & 
				\includegraphics[width=0.14\textwidth]{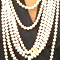} & 
				\includegraphics[width=0.14\textwidth]{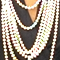} & 
				\includegraphics[width=0.14\textwidth]{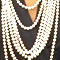} & 
				\includegraphics[width=0.14\textwidth]{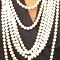} \\
				
				{} & 
				\includegraphics[width=0.14\textwidth]{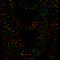} & 
				\includegraphics[width=0.14\textwidth]{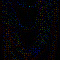} & 
				\includegraphics[width=0.14\textwidth]{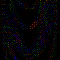} & 
				\includegraphics[width=0.14\textwidth]{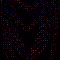} & 
				\includegraphics[width=0.14\textwidth]{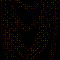} \\
				
				{} & CDM-3-Stage & LCNN-DD & JDNDMSR & Ours(-) & Ours \\
				Ground truth &  39.23/0.9846 & 39.54/0.9854 & 39.33/\textbf{0.9858} & 39.46/0.9847 & \textbf{39.68}/0.9854\\

			\end{tabular}
		} 
	\end{center}
	
	\caption{Results of the various comparisons between state-of-the-art and our method for noise-free demosaicking on image 18 of  Kodak. 
	}
	\label{demosaicking}
\end{figure*}


Figure~\ref{demosaicking} illustrates a challenging case in which existing algorithms always produce color distortions (in the necklace part), while the proposed algorithm presents no distortion. In order to better observe the reconstruction effect of the algorithm, we show the residual image between the reconstructed image generated by all the algorithms and the ground truth. It can be seen that the visual effect is consistent with the numerical evaluation.

\begin{sidewaystable}[!htpb]	
	\begin{center}
	\caption{Comparison of the results (PSNR/SSIM) between different denoising and demosaicking methods for five image sets.
	The best value is marked in {\textbf{bold}}, the second is marked in {\color{red}red}, and the third is marked in {\color{blue}blue}. The noise levels for which an algorithm doesn't work is indicated by "--" in the table. The algorithm noted by "*" which we don't obtain the source code, then the results (PSNR/SSIM) are taken from the article directly (the notation "x" in the table represents the unknown).}
	\label{DN_Kodak}
        \scriptsize
		\renewcommand\arraystretch{1. }
		\addtolength{\tabcolsep}{-2.9pt}
		\begin{tabular}{|c|c|c|c|c|c|c|c|c|c|c|}
			\hline
			\multicolumn{2}{|c|}{} & JCNN~\cite{gharbi2016deep} & C-RCNN~\cite{kokkinos2018deep} & LCNN-DD~\cite{Huang2018Lightweight}  & ADMM~\cite{Tan2017ADMM} & SGNet{\rm *}\cite{Liu_2020_CVPR} & JDNDMSR~\cite{xing2021end}  & 1.5CBM3D~\cite{jin2020review} & Ours(-) & Ours \\ 
			\multicolumn{2}{|c|}{\multirow{-2}{*}{Algorithm}} & PSNR/SSIM & PSNR/SSIM & PSNR/SSIM & PSNR/SSIM & PSNR/SSIM & PSNR/SSIM & PSNR/SSIM & PSNR/SSIM & PSNR/SSIM \\ \hline
			
			\multicolumn{2}{|c}{} & \multicolumn{9}{c|}{Kodak}  \\\hline
			& 3  & 37.95/0.9626 & 37.25/0.9587 & 37.36/0.9428 & 31.85/0.8813 & x & {\color{blue}38.93}/{\color{red}0.9678} & 38.73/0.9663 & {\color{red}38.97}/{\color{blue}0.9677} & \textbf{39.15}/\textbf{0.9685}     \\
			& 5  & 36.18/0.9446 & 35.28/0.9327 & 33.91/0.8704 & 31.81/0.8765 & x & {\color{blue}36.99}/{\color{blue}0.9510} & 36.57/0.9482 & {\color{red}37.01}/{\color{red}0.9511} & \textbf{37.12}/\textbf{0.9519} \\ 
			& 10 & 33.21/0.9007 & 30.94/0.8279 & 28.36/0.6710 & 31.22/0.8576 & x & {\color{blue}33.94}/{\color{blue}0.9115} & 33.34/0.9058 & {\color{red}33.97}/{\color{red}/0.9124} & \textbf{34.08}/\textbf{0.9143} \\  
			& 15 & 31.32/0.8586 & -- & 24.97/0.5201 & 30.30/0.8350 & x & {\color{blue}32.08}/{\color{blue}0.8752} & 31.44/0.8679 & {\color{red}32.12}/{\color{red}0.8780}  & \textbf{32.24}/\textbf{0.8807}   \\  
			& 20 & 29.91/0.8168 &-- & -- & 29.37/0.8115 & -- & {\color{blue}30.79}/{\color{blue}0.8430} & 30.13/0.8343 & {\color{red}30.89}/{\color{red}0.8487} & \textbf{31.01}/\textbf{0.8518}  \\ 
			& 40 & -- & -- & -- & 25.72/0.6797 & -- & -- & {\color{blue}26.88}/{\color{blue}0.7242} & {\color{red}28.00}/{\color{red}0.7621} & \textbf{28.13}/\textbf{0.7663}   \\ 
			\multirow{-6}{*}{$\sigma$} & 60 &  -- & -- & -- & 24.22/0.6256 & -- & -- & {\color{blue}24.80}/{\color{blue}0.6533} & {\color{red}26.46}/{\color{red}0.7074} & \textbf{26.58}/\textbf{0.7112}   \\ \hline 
			
			\multicolumn{2}{|c}{} & \multicolumn{9}{c|}{McMaster} \\ \hline
			& 3 & 36.44/0.9470 & 34.36/0.9222 & 35.98/0.9254 & 32.51/0.9048 & x & {\color{blue}37.38}/{\color{blue}0.9546} & 37.15/0.9509 & {\color{red}37.65}/{\color{red}0.9554} & \textbf{37.82}/\textbf{0.9567}       \\ 
			& 5 & 35.31/0.9338 & 33.18/0.8985 & 33.29/0.8584 & 32.46/0.8985 & x & {\color{blue}36.14}/{\color{blue}0.9421} & 35.54/0.9347 & {\color{red}36.29}/{\color{red}0.9423} & \textbf{36.41}/\textbf{0.9433} \\ 
			& 10 & 33.02/0.8972 & 29.66/0.7885 & 28.44/0.6740 & 31.64/0.8724 & x & {\color{blue}33.80}/{\color{blue}0.9126}  & 32.84/0.8954 & {\color{red}33.91}/{\color{red}0.9132} & \textbf{34.03}/\textbf{0.9152} \\  
			& 15 & 31.25/0.8564 & -- & 25.29/0.5309 &  30.46/0.8399  & x & {\color{blue}32.17}/{\color{blue}0.8858} & 31.03/0.8561 & {\color{red}32.25}/{\color{red}0.8867} & \textbf{32.40}/\textbf{0.8902}  \\ 
			& 20 & 29.79/0.8139 & -- & --  & 29.29/0.8068  & -- & {\color{blue}30.93}/{\color{blue}0.8608} & 29.66/0.8186 & {\color{red}31.06}/{\color{red}0.8635} & \textbf{31.21}/\textbf{0.8670} \\
			& 40 & -- & -- &-- & 25.12/0.6650  & -- & -- & {\color{blue}25.90}/{\color{blue}0.6971} & {\color{red}28.04}/{\color{red}0.7902}  & \textbf{28.21}/\textbf{0.7962}   \\ 
			\multirow{-6}{*}{$\sigma$} & 60 & -- & -- & -- & 22.92/0.5957  & -- & -- & {\color{blue}23.33}/{\color{blue}0.6155} & {\color{red}26.28}/{\color{red}0.7369} & \textbf{26.45}/\textbf{0.7422}   \\ \hline		
			
			\multicolumn{2}{|c}{} & \multicolumn{9}{c|}{WED + Flickr} \\ \hline
			& 3 & 36.28/0.9592 & 35.19/0.9486 & 35.92/0.9372 & 31.35/0.9060 & x & 37.32/{\color{blue}0.9663} & {\color{blue}37.38}/0.9646 & {\color{red}37.72}/{\color{red}0.9676} & \textbf{37.92}/\textbf{0.9685} \\
			& 5 & 35.07/0.9448 & 33.77/0.9241 & 33.09/0.8660 & 31.32/0.9003 & x & {\color{blue}35.98}/{\color{blue}0.9542} & 35.69/0.9485 & {\color{red}36.27}/{\color{red}0.9554} & \textbf{36.40}/\textbf{0.9563}  \\   
			& 10 & 32.70/0.9081 & 30.38/0.8393 & 28.15/0.6758 & 30.72/0.8808  & x & {\color{blue}33.56}/{\color{blue}0.9258} & 32.85/0.9108 & {\color{red}33.76}/{\color{red}0.9276} & \textbf{33.88}/\textbf{0.9294}   \\  
			& 15 & 30.96/0.8719 & -- & 24.98/0.5363 & 29.78/0.8580  & x & {\color{blue}31.93}/{\color{blue}0.8998} & 31.00/0.8778 & {\color{red}32.07}/{\color{red}0.9022} & \textbf{32.20}/\textbf{0.9048}   \\ 
			& 20 & 29.53/0.8354 & -- & -- & 28.78/0.8341  & -- & {\color{blue}30.71}/{\color{blue}0.8762} & 29.63/0.8484 & {\color{red}30.86}/{\color{red}0.8800} & \textbf{31.00}/\textbf{0.8832}  \\ 
			& 40 & -- & -- & -- & 24.94/0.7167  & --  & -- & {\color{blue}25.93}/{\color{blue}0.7523} & {\color{red}27.84}/{\color{red}0.8083} & \textbf{27.98}/\textbf{0.8127}       \\ 
			\multirow{-6}{*}{$\sigma$} & 60 & -- & -- & -- & 22.84/0.6625 & -- & -- & {\color{blue}23.40}/{\color{blue}0.6832} & {\color{red}26.10}/{\color{red}0.7565} & \textbf{26.23}/\textbf{0.7613}       \\  \hline
			
			\multicolumn{2}{|c}{} & \multicolumn{9}{c|}{Urban 100} \\ \hline 
			& 3 & 34.87/0.9680 & 34.98/0.9680 & 35.25/0.9560 & 28.53/0.9010 & x & 36.47/{\color{blue}0.9749} & {\color{blue}36.70}/0.9741 & {\color{red}37.07}/{\color{red}0.9768} & \textbf{37.40}/\textbf{0.9779}       \\
			& 5 & 33.69/0.9569 & 33.47/0.9526 & 32.72/0.9101 & 28.71/0.8987 & 34.54/0.9533 & {\color{blue}35.11}/{\color{blue}0.9665} & 34.89/0.9621 & {\color{red}35.52}/{\color{red}0.9683} & \textbf{35.77}/\textbf{0.9694}    \\ 
			& 10 & 31.26/0.9248 & 29.90/0.8895 & 27.97/0.7776 & 28.67/0.8864 & 32.14/0.9229 & {\color{blue}32.57}/{\color{blue}0.9438}  & 31.86/0.9307 & {\color{red}32.81}/{\color{red}0.9459} & \textbf{33.04}/\textbf{0.9482}   \\  
			& 15 & 29.45/0.8912 & -- & 24.82/0.6692 & 28.08/0.8681 & 30.37/0.8923 & {\color{blue}30.79}/{\color{blue}0.9206}  & 29.96/0.9018 & {\color{red}30.96}/{\color{red}0.9233} & \textbf{31.21}/\textbf{0.9268}   \\ 
			& 20 & 28.00/0.8552 & -- & -- & 27.26/0.8463  & -- & {\color{blue}29.44}/{\color{blue}0.8972} & 28.58/0.8744 & {\color{red}29.59}/{\color{red}0.9012} & \textbf{29.86}/\textbf{0.9060}  \\
			& 40 & -- & -- & -- & 23.60/0.7209 & -- & -- & {\color{blue}24.99}/{\color{blue}0.7726} & {\color{red}26.17}/{\color{red}0.8182} & \textbf{26.47}/\textbf{0.8277}     \\ 
			\multirow{-6}{*}{$\sigma$} & 60 & -- & -- & -- & 22.05/0.6628 & -- & -- & {\color{blue}22.56}/{\color{blue}0.6833} & {\color{red}24.19}/{\color{red}0.7476} & \textbf{24.45}/\textbf{0.7582}  \\ \hline
			
			\multicolumn{2}{|c}{} & \multicolumn{9}{c|}{MIT moiré} \\ \hline 
			& 3 & 33.66/0.9331 & 31.69/0.8853 & 33.08/0.9094 & 28.44/0.8296 & x & 33.97/0.9216 & {\color{blue}34.69}/{\color{blue}0.9360} & {\color{red}34.84}/{\color{red}0.9404} & \textbf{35.28}/\textbf{0.9427}   \\
			& 5 & 32.57/0.9172 & 30.68/0.8686  & 31.27/0.8699 & 28.48/0.8237 & 32.15/0.9043 & 32.84/0.9094 & {\color{blue}33.21}/{\color{blue}0.9145} & {\color{red}33.52}/{\color{red}0.9266} & \textbf{33.81}/\textbf{0.9289}   \\ 
			& 10 & 30.39/0.8724 & 28.08/0.8008 & 27.33/0.7465 & 28.19/0.8031 & 30.09/0.8619 & {\color{blue}30.73}/{\color{blue}0.8759} & 30.71/0.8701 & {\color{red}31.19}/{\color{red}0.8902} & \textbf{31.40}/\textbf{0.8939}   \\  
			& 15 & 28.81/0.8283 & -- & 24.46/0.6347 & 27.55/0.7801 & 28.60/0.8188 & {\color{blue}29.28}/{\color{blue}0.8427} & 29.12/0.8344 & {\color{red}29.60}/{\color{red}0.8544} & \textbf{29.83}/\textbf{0.8597}   \\ 
			& 20 & 27.57/0.7842 & -- & -- & 26.82/0.7564 & -- & {\color{blue}28.21}/0.8107 & 27.95/{\color{blue}0.8019} & {\color{red}28.47}/{\color{red}0.8213} & \textbf{28.71}/\textbf{0.8288}  \\
			& 40 & -- & -- & -- & 23.76/0.6387 & -- & -- & {\color{blue}24.96}/{\color{blue}0.6882} & {\color{red}25.77}/{\color{red}0.7176} & \textbf{26.01}/\textbf{0.7279} \\ 
			\multirow{-6}{*}{$\sigma$} & 60 & -- & -- & -- & 22.43/0.5806 & -- & -- & {\color{blue}22.93}/{\color{blue}0.5946} & {\color{red}24.25}/{\color{red}0.6421} & \textbf{24.46}/\textbf{0.6535} \\ \hline
		\end{tabular}
	\end{center}

\end{sidewaystable}

\paragraph{Joint demosaicking and denoising}
For the task of demosaicking and denoising of noisy CFA images, we compared with the joint demosaicking and denoising algorithm using ADMM by~\cite{Tan2017ADMM}. The joint demosaicking and denoising algorithms
based on deep learning proposed in~\cite{gharbi2016deep} (JCNN), in~\cite{kokkinos2018deep} (C-RCNN), in~\cite{Huang2018Lightweight} (LCNN-DD), in~\cite{Liu_2020_CVPR} (SGNet\footnote{Since we didn't obtain the source code of the algorithm SGNet \cite{Liu_2020_CVPR}, 
the PSNR and SSIM values of the algorithm were taken from the article directly.}) and in~\cite{xing2021end} (JDNDMSR). 
We also considered our proposed demosaicking network combined with CBM3D for denoising~\cite{Dabov2007CBM3D}, and following the suggestion of Jin {\em et al.}\cite{jin2020review}, the CBM3D denoising parameter was set to 1.5 times the original $\sigma$ value (denoted 1.5CBM3D).
Table~\ref{DN_Kodak} summarizes the performance comparison of all algorithms.
It can be seen that our algorithm performs better than other state-of-the-art algorithms.

\begin{figure*}[t]
\begin{center}
    \tiny
	\renewcommand{\arraystretch}{0.5} \addtolength{\tabcolsep}{-5pt} {%
		\begin{tabular}{ccccccc}
				\multirow{2}{*}[20pt]{\includegraphics[width=0.16\textwidth]{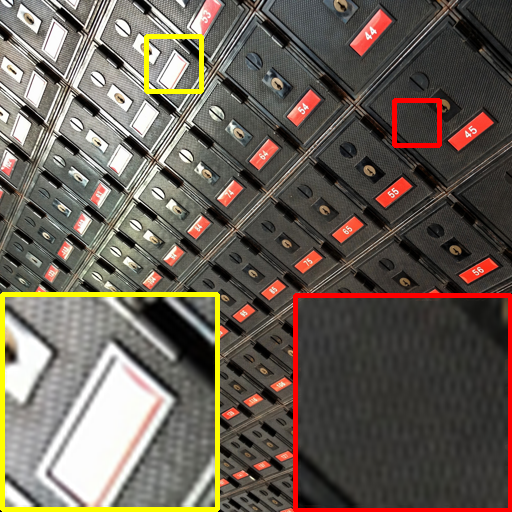}} &
				\includegraphics[width=0.135\textwidth]{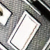} &
				\includegraphics[width=0.135\textwidth]{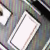} &
				\includegraphics[width=0.135\textwidth]{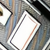} &
				\includegraphics[width=0.135\textwidth]{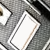} &
				\includegraphics[width=0.135\textwidth]{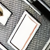} &
				\includegraphics[width=0.135\textwidth]{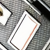} \\
				
				{} &
				\includegraphics[width=0.135\textwidth]{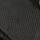} &
				\includegraphics[width=0.135\textwidth]{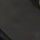} &
				\includegraphics[width=0.135\textwidth]{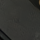} &
				\includegraphics[width=0.135\textwidth]{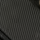} &
				\includegraphics[width=0.135\textwidth]{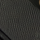} &
				\includegraphics[width=0.135\textwidth]{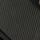} \\
				
				{} & JCNN & ADMM & JDNDMSR & 1.5CBM3D & Ours(-) & Ours \\
			    Ground truth & 30.40/0.9149 & 26.70/0.8391 & 31.48/0.9240 & 30.95/0.9317 & 31.88/0.9345 & \textbf{32.12}/\textbf{0.9386} \\
				
			\end{tabular}
		} 
	\end{center}
	
	\caption{Comparison between state-of-the-art algorithms and our method for demosaicking and denoising in image 6 of the Urban dataset with noise $\sigma=10$.}
	\label{sigma_10}
\end{figure*}

\begin{figure*}[t]
\begin{center}
    \tiny
	\renewcommand{\arraystretch}{0.5} \addtolength{\tabcolsep}{-5pt} {%
		\begin{tabular}{ccccccc}
				\multirow{2}{*}[20pt]{\includegraphics[width=0.16\textwidth]{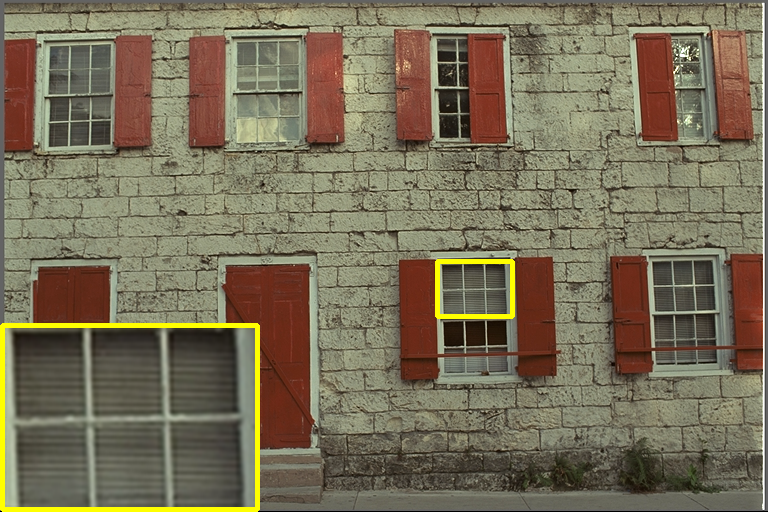}} &
				\includegraphics[width=0.135\textwidth]{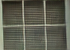} &
				\includegraphics[width=0.135\textwidth]{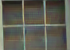} &
				\includegraphics[width=0.135\textwidth]{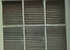} &
				\includegraphics[width=0.135\textwidth]{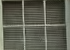} &
				\includegraphics[width=0.135\textwidth]{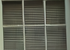} &
				\includegraphics[width=0.135\textwidth]{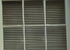} \\
				
				{} &
				\includegraphics[width=0.135\textwidth]{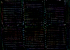} &
				\includegraphics[width=0.135\textwidth]{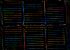} &
				\includegraphics[width=0.135\textwidth]{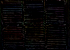} &
				\includegraphics[width=0.135\textwidth]{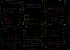} &
				\includegraphics[width=0.135\textwidth]{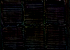} &
				\includegraphics[width=0.135\textwidth]{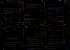} \\
				
				{} & JCNN & ADMM & JDNDMSR & 1.5CBM3D & Ours(-) & Ours \\
			    Ground truth & 29.14/0.8510 & 27.98/0.7951 & 29.58/0.8626  & 29.15/0.8564  & 29.55/0.8635 & \textbf{29.63}/\textbf{0.8660} \\
				
			\end{tabular}
		} 
	\end{center}
	
	\caption{Comparison between state-of-the-art algorithms and our method for demosaicking and denoising in image 1 of the Kodak dataset with noise $\sigma=15$.}
	\label{sigma_20}
\end{figure*}

\begin{figure*}[!ht]
	\begin{center}
	    \tiny
		\renewcommand{\arraystretch}{0.5} \addtolength{\tabcolsep}{-5pt} {%
			\begin{tabular}{ccccccc}
				\multirow{2}{*}[40pt]{\includegraphics[width=0.16\textwidth]{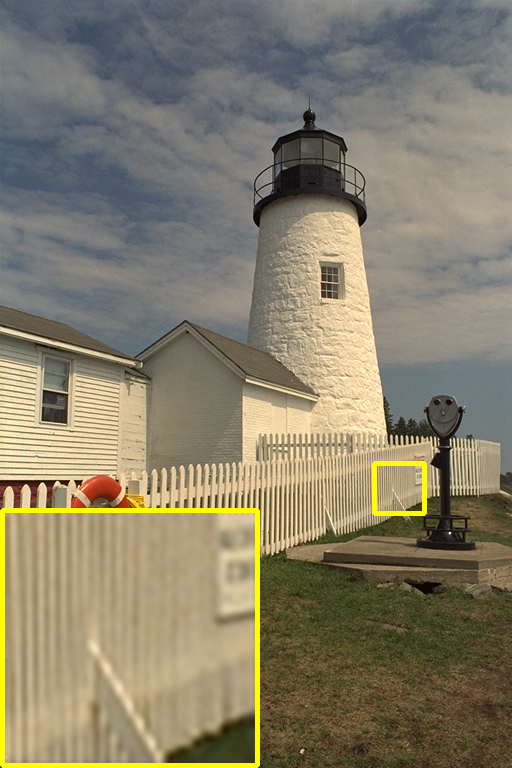}} &
				\includegraphics[width=0.135\textwidth]{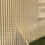} &
				\includegraphics[width=0.135\textwidth]{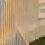} &
				\includegraphics[width=0.135\textwidth]{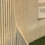} &
				\includegraphics[width=0.135\textwidth]{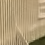} &
				\includegraphics[width=0.135\textwidth]{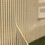} &
				\includegraphics[width=0.135\textwidth]{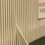} 
				\\
				
				{} &
				\includegraphics[width=0.135\textwidth]{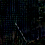} &
				\includegraphics[width=0.135\textwidth]{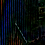} &
				\includegraphics[width=0.135\textwidth]{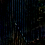} &
				\includegraphics[width=0.135\textwidth]{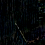} &
				\includegraphics[width=0.135\textwidth]{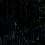} &
				\includegraphics[width=0.135\textwidth]{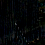} 
				\\
				
				{} & JCNN & ADMM & JDNDMSR & 1.5CBM3D & Ours(-) & Ours \\

				Ground truth & 30.07/0.8092 & 29.07/0.7994 & 30.78/0.8318 & 30.48/0.8299 & 30.91/0.8419 & \textbf{31.03}/\textbf{0.8442} \\
			\end{tabular}
		} 
	\end{center}
	
	\caption{Comparison between state-of-the-art algorithms and our method for demosaicking and denoising in image 1 of the Kodak dataset with noise $\sigma=20$.}
	\label{sigma_20_1}
\end{figure*}

\begin{figure*}[!htpb]
	\begin{center}
	    \tiny
		\renewcommand{\arraystretch}{0.5} \addtolength{\tabcolsep}{-5pt} {%
			\begin{tabular}{ccccccc}
				
				\multirow{2}{*}[30pt]{\includegraphics[width=0.16\textwidth]{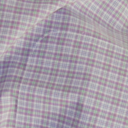}} &
				\includegraphics[width=0.135\textwidth]{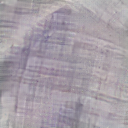} & 
				\includegraphics[width=0.135\textwidth]{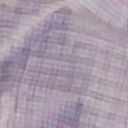} & 
				\includegraphics[width=0.135\textwidth]{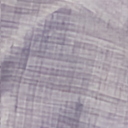} & 
				\includegraphics[width=0.135\textwidth]{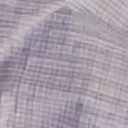} & 
				\includegraphics[width=0.135\textwidth]{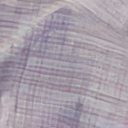} & 
				\includegraphics[width=0.135\textwidth]{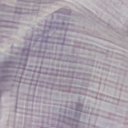} \\
				
				 &
				\includegraphics[width=0.135\textwidth]{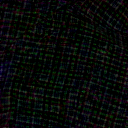} & 
				\includegraphics[width=0.135\textwidth]{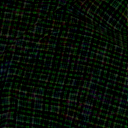} & 
				\includegraphics[width=0.135\textwidth]{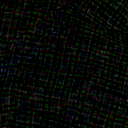} & 
				\includegraphics[width=0.135\textwidth]{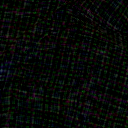} & 
				\includegraphics[width=0.135\textwidth]{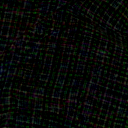} &
				\includegraphics[width=0.135\textwidth]{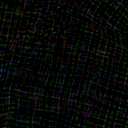} \\
				
				Ground truth & JCNN & ADMM & JDNDMSR & 1.5CBM3D & Ours(-) & Ours \\
				{} & 28.81/0.7077 & 28.90/0.7221 & 29.79/0.7796 & 29.48/0.7714 & 29.75/0.7740 & \textbf{30.35}/\textbf{0.8086} \\
				
			\end{tabular}
		} 
		
	\end{center}
	\caption{Comparison between state-of-the-art algorithms and our method for demosaicking and denoising in image 585 of the MIT moiré with noise $\sigma=20$.}
	\label{sigma_20moire}
\end{figure*}

Figure~\ref{sigma_10}-\ref{sigma_20moire} show the comparison of  visual effects and image quality between the state-of-the-art and our proposed method.
As can be seen in Figure~\ref{sigma_10} and~\ref{sigma_20moire}, our restored images show a more distinct image texture and fine detail.
Figure~\ref{sigma_20} illustrates that on the fence:  our restored image is more pleasant and has fewer color distortions and checkerboard artifacts. 
We also note that CBM3D + our proposed demosaicking also outperforms the state-of-the-art for both quantitative and visual quality.

\begin{table}[!t]
	\begin{center}
	\caption{Comparison of the results (PSNR/SSIM) between different flexible joint demosaicking and denoising methods in the interval of the noise level $\sigma \in (0,20]$ for five image sets. The best value is marked in {\textbf{bold}}, the second is marked in {\color{red}red}.}
	\label{F}
	    \scriptsize
		\renewcommand\arraystretch{1.2}
		\addtolength{\tabcolsep}{-3.7pt}
		\begin{tabular}{|c|l|c|c|c|c|}
			\hline
            $\sigma$  & Dataset  & JCNN  & JDNDMSR & Ours(-)-F & Ours-F \\ \hline
            \multirow{5}{*}{10} & Kodak & 33.21/0.9007 & {\color{red}33.94}/0.9115 & 33.92/{\color{red}0.9116} & \textbf{34.03}/\textbf{0.9129} \\
            &  McMaster & 33.02/0.8972 & 33.80/{\color{red}0.9126} & {\color{red}33.85}/0.9121 & \textbf{33.97}/\textbf{0.9142} \\
            &  WED + Flickr & 32.70/0.9081 & 33.56/0.9258 & {\color{red}33.70}/{\color{red}0.9269} & \textbf{33.83}/\textbf{0.9284} \\
            &  Urban 100 & 31.26/0.9248 & 32.57/0.9438 & {\color{red}32.75}/{\color{red}0.9454} & \textbf{32.95}/\textbf{0.9471} \\
            &  MIT moiré & 30.39/0.8724 & 30.73/0.8759 & {\color{red}31.09}/{\color{red}0.8884} & \textbf{31.31}/\textbf{0.8919} \\ \hline
            
            \multirow{5}{*}{20} & Kodak & 29.91/0.8168 & {\color{red}30.79}/0.8430 & 30.75/{\color{red}0.8441} & \textbf{30.86}/\textbf{0.8465} \\
            & McMaster     & 29.79/0.8139 & {\color{red}30.93}/{\color{red}0.8608} & 30.88/0.8571 & \textbf{31.02}/\textbf{0.8609} \\
            & WED + Flickr & 29.53/0.8354 & {\color{red}30.71}/{\color{red}0.8762} & {\color{red}30.71}/0.8750 & \textbf{30.84}/\textbf{0.8780} \\
            & Urban 100    & 28.00/0.8552 & {\color{red}29.44}/{\color{red}0.8972} & 29.39/0.8963 & \textbf{29.61}/\textbf{0.9002} \\
            & MIT moiré    & 27.57/0.7842 & 28.21/0.8107 & {\color{red}28.28}/{\color{red} 0.8141} & \textbf{28.49}/\textbf{0.8207} \\ \hline
		\end{tabular}
	\end{center}
\end{table}

\paragraph{Noise level flexible joint demosaicking and denoising}
Referring to \cite{gharbi2016deep, Liu_2020_CVPR, xing2021end}, the noise level map was introduced in the denoising stage to flexibly handle 
the noise of a certain range of noise levels
($\sigma \in (0,20]$). 
The corresponding PSNR and SSIM values are shown in Table~\ref{F}.
One can observe that the proposed method is superior to JCNN~\cite{gharbi2016deep} and JDNDMSR~\cite{xing2021end} for all five image databases.    
Our lightweight method is very competitive with JDNDMSR~\cite{xing2021end} and outperforms JCNN~\cite{gharbi2016deep}.

\subsection{Results on real image datasets}
Since the raw data is represented in the linear RGB color space (ie, without gamma transformation), inspired by~\cite{Guo2021Joint, Brooks2019Unprocessing}, we used the unprocessing algorithm~\cite{Brooks2019Unprocessing} to convert the training data to linear RGB data and fine-tune the proposed algorithm. 
We evaluated the proposed algorithm on real images from the Darmstadt Noise Dataset (DND)~\cite{2017DND}. 
Since there is no ground truth for these real world images, we decided to use the qualitative
 natural image quality evaluator (NIQE)~\cite{Mittal2013Making} to evaluate the perceptual quality of the reconstructed images. 
The  only input of NIQE is  the restored image.
 Lower NIQE scores mean higher image quality.
The   NIQE scores of the restored results for real images were compared in linear RGB and sRGB color spaces for the various algorithms and are shown in Table~\ref{Tab_DND}. The NIQE scores of our method and of its lightweight version are much lower than that of JCNN. Figure~\ref{DND} shows the restored images of the various algorithms in the sRGB space  providing a visual quality confirmation of these measurements.
A key part of each image in a red box is zoomed in  and placed on the right side to make comparison easier.
One can see that our proposed method restores better the textures and  suppresses more noise than JCNN. Taking the first column of Figure~\ref{DND} for example, the restored image of JCNN can't  recover the top left curve, which is broken in the middle.

%
\begin{table}[t]
	\caption{NIQE comparison between our proposed method and JCNN in DND dataset.}
	\label{Tab_DND}
	\begin{center}
	    \scriptsize
		\renewcommand\arraystretch{1.2}
		\begin{tabular}{|l|c|c|c|}
			\hline
		       & JCNN    & Ours(-) & Ours \\ \hline
		linRGB & 13.2701 & 8.8395 & \textbf{4.3157} \\ 
		sRGB   & 12.2354 & 7.8584 & \textbf{3.8507} \\ \hline
		\end{tabular}
	\end{center}
\end{table}

\begin{figure}[!t]
\begin{center}
    \tiny
	\renewcommand{\arraystretch}{0.5} \addtolength{\tabcolsep}{-6pt} {%
		\begin{tabular}{ccccccccccccc}
			\includegraphics[width=0.119\textwidth]{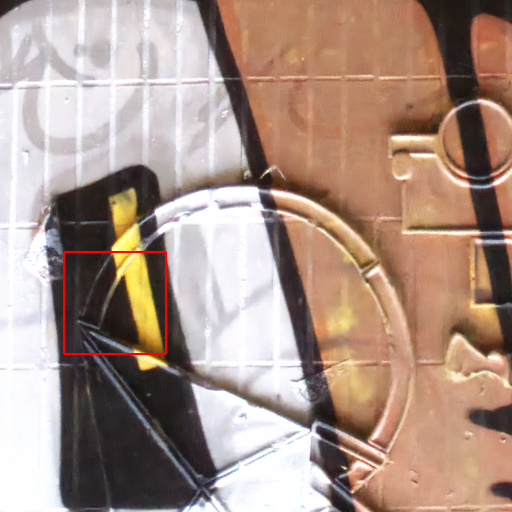} &
			\includegraphics[width=0.119\textwidth]{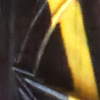} &$\ $ &
			\includegraphics[width=0.119\textwidth]{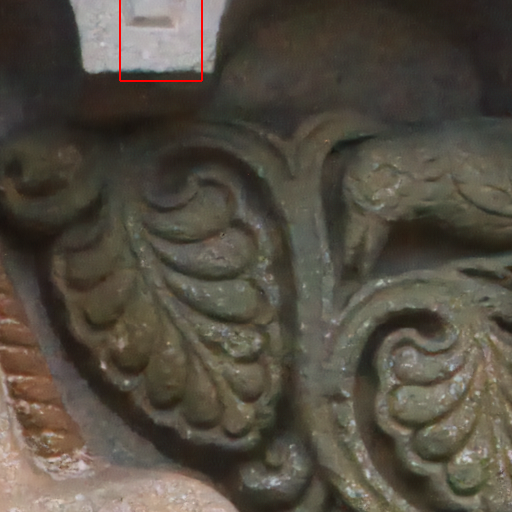} &
			\includegraphics[width=0.119\textwidth]{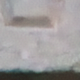} &$\ $ &
			\includegraphics[width=0.119\textwidth]{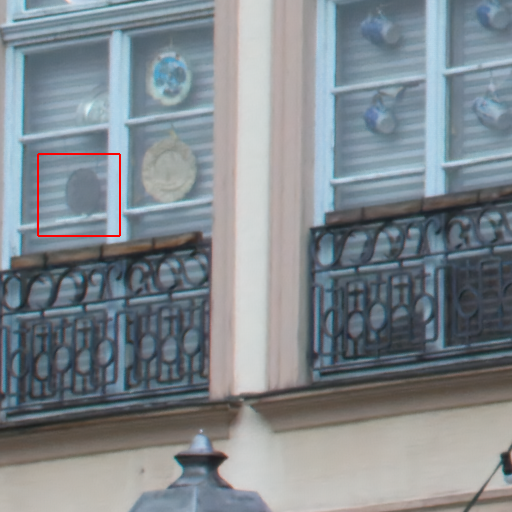} &
			\includegraphics[width=0.119\textwidth]{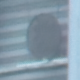} &$\ $ &
			\includegraphics[width=0.119\textwidth]{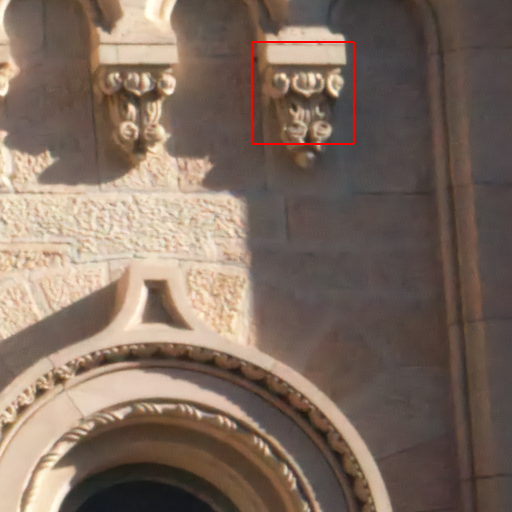} &
			\includegraphics[width=0.119\textwidth]{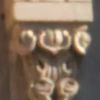} \\
			\multicolumn{2}{c}{JCNN NIQE=3.5533} &$\ $ & \multicolumn{2}{c}{NIQE=4.3134} &$\ $ & \multicolumn{2}{c}{NIQE=5.1885} &$\ $ & \multicolumn{2}{c}{NIQE=4.5778} \\
			
			\includegraphics[width=0.119\textwidth]{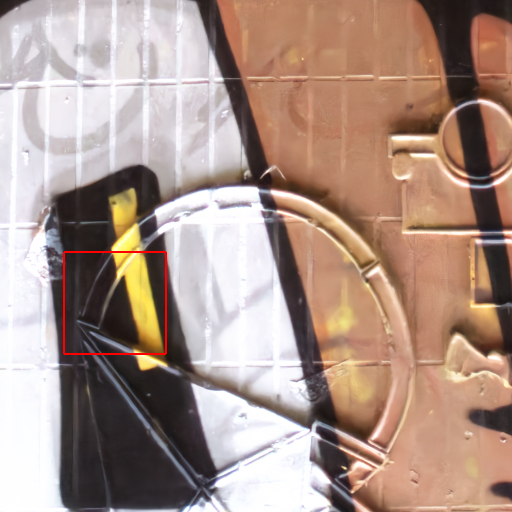} &
			\includegraphics[width=0.119\textwidth]{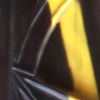} &$\ $ &
			\includegraphics[width=0.119\textwidth]{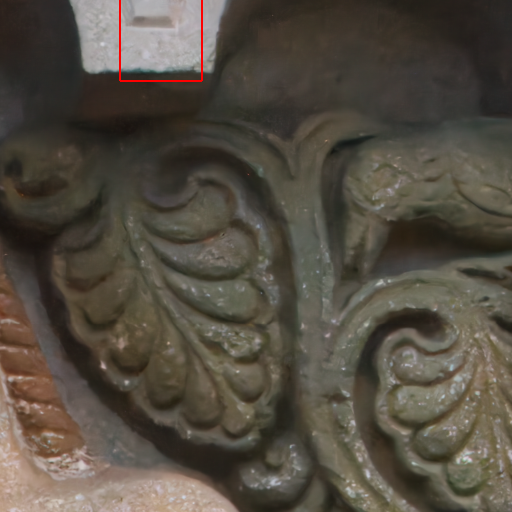} &
			\includegraphics[width=0.119\textwidth]{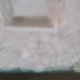} &$\ $ &
			\includegraphics[width=0.119\textwidth]{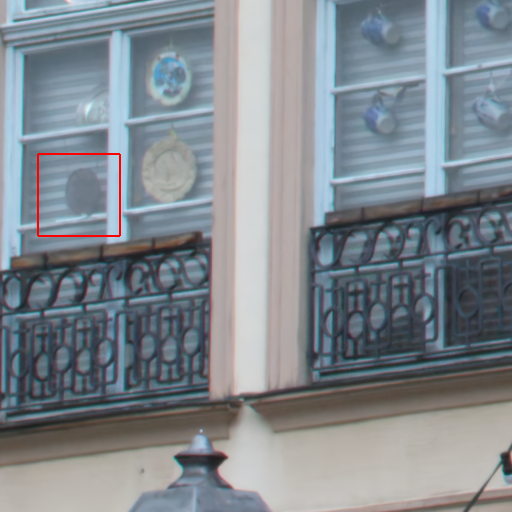} &
			\includegraphics[width=0.119\textwidth]{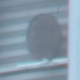} &$\ $ &
			\includegraphics[width=0.119\textwidth]{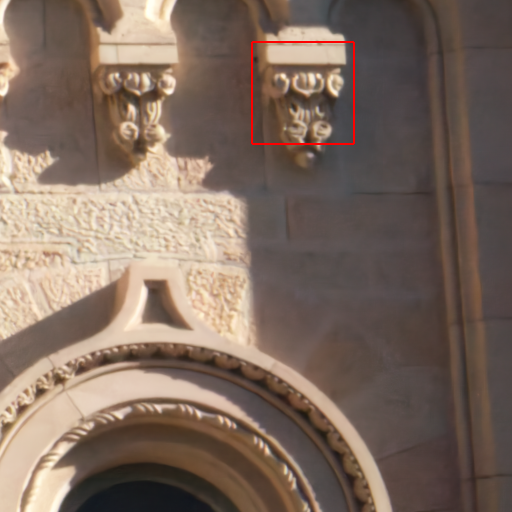} &
			\includegraphics[width=0.119\textwidth]{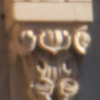} \\
			\multicolumn{2}{c}{Ours(-) NIQE=3.2810} &$\ $ & \multicolumn{2}{c}{NIQE=3.5681} &$\ $ & \multicolumn{2}{c}{NIQE=4.9610} &$\ $ & \multicolumn{2}{c}{NIQE=4.2286}  \\
			
			\includegraphics[width=0.119\textwidth]{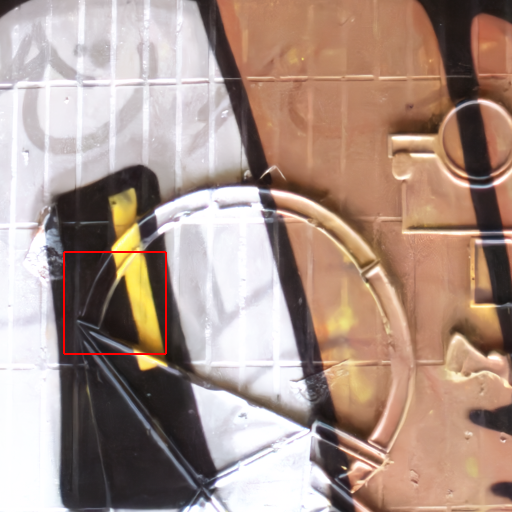} &
			\includegraphics[width=0.119\textwidth]{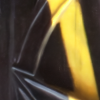} &$\ $ &
			\includegraphics[width=0.119\textwidth]{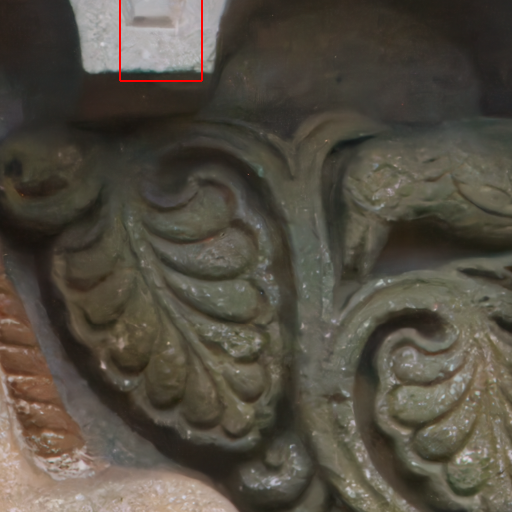} &
			\includegraphics[width=0.119\textwidth]{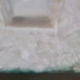} &$\ $ &
			\includegraphics[width=0.119\textwidth]{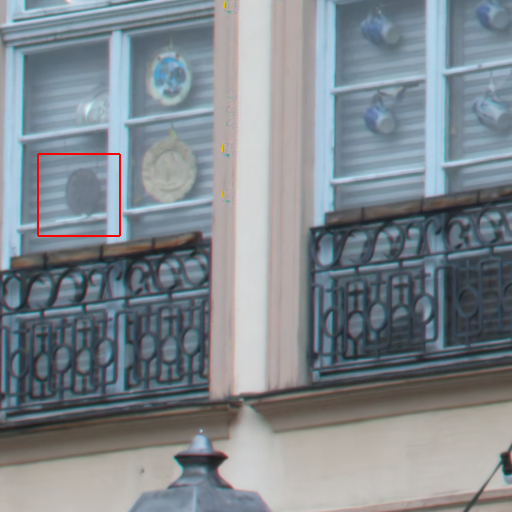} &
			\includegraphics[width=0.119\textwidth]{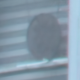} &$\ $ &
			\includegraphics[width=0.119\textwidth]{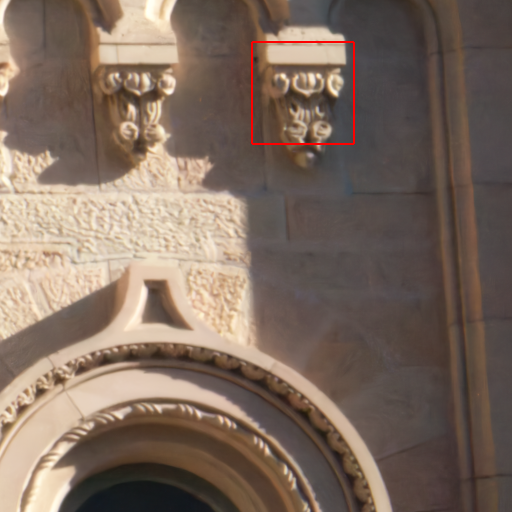} &
			\includegraphics[width=0.119\textwidth]{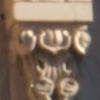} \\
			\multicolumn{2}{c}{Ours NIQE=3.1433}&$\ $ & \multicolumn{2}{c}{NIQE=3.3624} &$\ $ & \multicolumn{2}{c}{NIQE=4.3052} &$\ $ & \multicolumn{2}{c}{NIQE=3.8501} \\

		\end{tabular}
		}
\end{center}
\caption{Comparison of JCNN with our method for demosaicking and denoising in real images of DND. Each group of images consists of the whole image and a part of the image. The image on the left is the whole image, and the image on the right is the zoomed in image of the part in the red box on the left.}
\label{DND}
\end{figure}

\subsection{Ablation study and running time}

\paragraph{Architecture choices, ablation study}
Our ablation experiments trained and compared the following models: 
\begin{itemize}
 \item (a) Using  GBTF~\cite{Pekkucuksen2010gbtf} for preprocessing, while the demosaicking network is built using 16 classic Conv-BN-ReLU blocks (consistent with our network parameters). 
 \item (b) Using  GBTF~\cite{Pekkucuksen2010gbtf} for preprocessing, while the demosaicking network is built using 8 Resblocks (consistent with our network parameters). 
 \item (c) Using HA~\cite{hamilton1997adaptive} for preprocessing, while the network uses our proposed Inception block.
 \item (d) Using bilinear interpolation for preprocessing, while the network uses our proposed Inception block.
\end{itemize}
The performance of the above four cases on the five datasets is shown in Table~\ref{Ablation_all}~(A).
As can be seen from the table, good results can also be obtained using bilinear interpolation, but  GBTF  is a better choice when working with textured images.
The table also shows that   GBTF  for preprocessing and using Inception blocks are more effective for image demosaicking.

\begin{sidewaystable}[!htpb]
	\caption{Ablation study. Sub-table A is the choice of network structure. Sub-table B shows the comparison of two-stage training with end-to-end training.}
	\label{Ablation_all}
	\scriptsize
	\renewcommand\arraystretch{1.2}

	\begin{tabular}{|l|c|c|c|c|c|c|}
		\hline
		\multirow{2}{*}{Method} & Kodak & McMaster & WED+Flickr  & Urban100 & MIT moiré & Average \\ 
		{}  & PSNR/SSIM & PSNR/SSIM & PSNR/SSIM & PSNR/SSIM & PSNR/SSIM & PSNR/SSIM    \\ \hline
		
	             \multicolumn{7}{|c|}{A. Demosaic} \\\hline
		GBTF+Conv       & 42.38/0.9886 & 39.28/0.9707 & 39.79/0.9818 & 38.78/0.9851 & 35.78/0.9503 & 39.20/0.9753 \\ 
        GBTF+Resblock   & 42.34/0.9884 & 39.32/0.9712 & 39.83/0.9820 & 38.81/0.9852 & 35.78/0.9501 & 39.22/0.9754 \\
		HA+Inception    & 42.14/0.9877 & 39.28/0.9707 & 39.71/0.9816 & 38.62/0.9849 & 35.72/0.9501 & 39.09/0.9750 \\
		Billinear+Inception  & 42.60/0.9889 & \textbf{39.62}/\textbf{0.9727} & 40.17/0.9830 & 39.21/0.9862 & 36.35/0.9536 & 39.59/0.9769 \\  
		Ours(-)  & 42.49/0.9888 & 39.25/0.9702 & 39.84/0.9820 & 38.88/0.9852 & 35.97/0.9516 & 39.29/0.9756 \\ 
		Ours     & \textbf{42.76}/\textbf{0.9893} & 39.61/0.9725 & \textbf{40.22}/\textbf{0.9831} & \textbf{39.52}/\textbf{0.9864} & \textbf{36.53}/\textbf{0.9533} & \textbf{39.73}/\textbf{0.9769}\\ \hline
		
	   \multicolumn{7}{|c|}{B. End-to-End Joint demosaicking and Denosing  ($\sigma=20$)}  \\	\hline
	   End-to-end training(-) & 30.71/0.8436 & 30.87/0.8587 & 30.66/0.8750 & 29.20/0.8937 & 28.11/0.8090 & 29.91/0.8560 \\
	   Two-Stage training(-) & 30.89/0.8487 & 31.06/0.8635 & 30.86/\textbf{0.8800} & 29.59/\textbf{0.9012} & 28.47/0.8213 & 30.17/\textbf{0.8629} \\
	   End-to-end FT of Two-stage(-) & \textbf{30.90}/\textbf{0.8489} & \textbf{31.08}/\textbf{0.8637} & \textbf{30.87}/0.8796 & \textbf{29.61}/0.9008 & \textbf{28.49}/\textbf{0.8214} & \textbf{30.19}/\textbf{0.8629} \\
	   \hline
	   End-to-end training & 30.99/0.8514 & 31.17/0.8664 & 30.97/0.8828 & 29.81/0.9051 & 28.61/0.8263 & 30.31/0.8664  \\
	   Two-Stage training & 31.01/0.8518 & 31.21/0.8670 & \textbf{31.00}/\textbf{0.8832} & 29.86/\textbf{0.9060} & 28.71/0.8288 &  30.36/0.8674 \\
    	   End-to-end FT of Two-stage & \textbf{31.02}/\textbf{0.8520} & \textbf{31.22}/\textbf{0.8671} & \textbf{31.00}/0.8829 & \textbf{29.87}/0.9059 & \textbf{28.73}/\textbf{0.8294} & \textbf{30.37}/\textbf{0.8675} \\ \hline
	     
	\end{tabular}
\end{sidewaystable}

\begin{figure*}[t]
	\centering
	\addtolength{\tabcolsep}{-6pt}
	{\fontsize{8pt}{\baselineskip}\selectfont 
	\begin{tabular}{cc}
		\includegraphics[width=0.45\textwidth, trim={20 3 49 27}, clip]{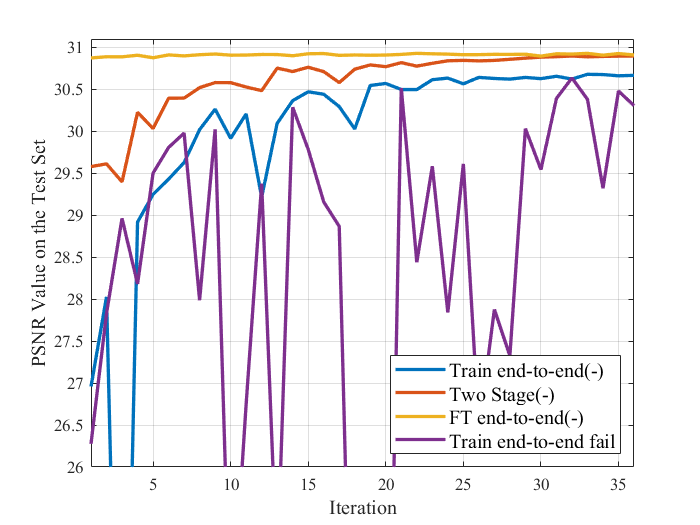} &
		\includegraphics[width=0.45\textwidth, trim={20 3 49 27}, clip]{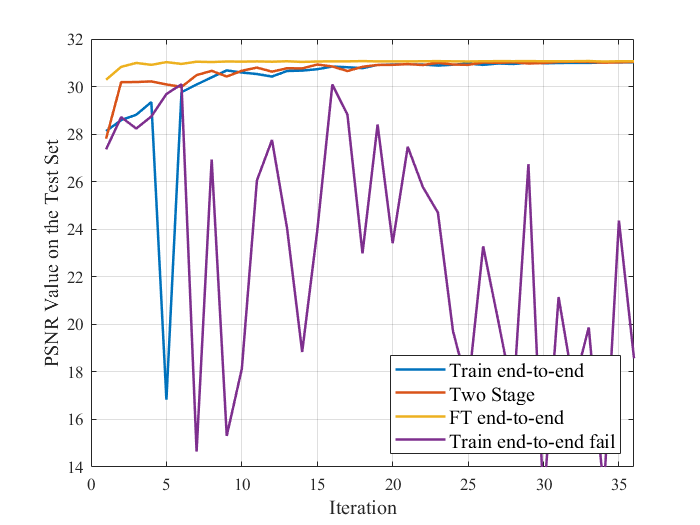} \\
		(a) Lightweight(-) & (b) Normal \\
		\includegraphics[width=0.45\textwidth, trim={20 3 49 27}, clip]{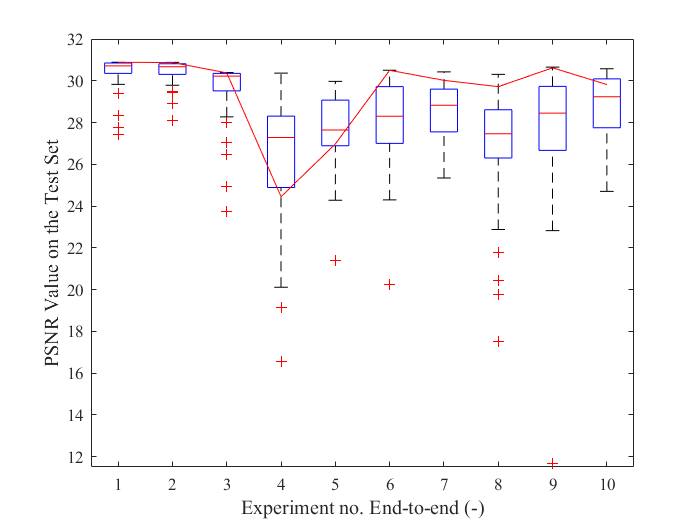} &
		\includegraphics[width=0.45\textwidth, trim={20 3 49 27}, clip]{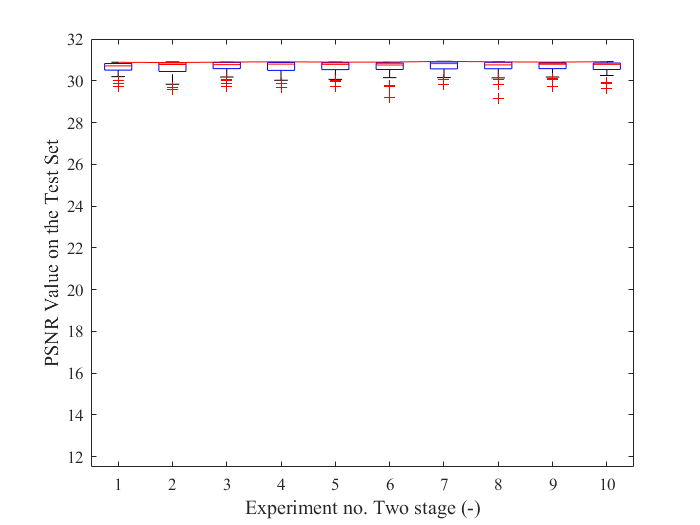} \\
		 \multicolumn{2}{c}{(c) Training ten times each} \\
	\end{tabular}
	}
	\caption{Plots (a) and (b) compare the performance of different training strategies along the training iterations. We compare the end-to-end training, the two-stage training and  finetuning after the two-stage training.
	We also report the evolution of a failed end-to-end training (purple curve), which were obtained with the same parameters as the blue curve. In (c), the end-to-end network and the two-stage network were each trained 10 times (for this experiment we used only the lightweight architecture). The unstable behavior (as in the purple curve) was observed in eight out of ten end-to-end  trainings, while the the two stage training never exhibited such behavior. The red curve marks the PSNR reached at the end of each training.}
	\label{end-to-end}
\end{figure*}

\paragraph{Two-stage vs. end-to-end training}
To verify the importance of the two-stage training we compared it with a joint demosaicking and denoising network trained  end-to-end.
For this experiment we set the noise level to $\sigma=20$.
Table~\ref{Ablation_all}~(B) shows the difference between both strategies. We can see that the end-to-end training of the networks  (with equivalent capacity)  is not as effective as the two-stage  training. This highlights the importance of training  first the demosaicking network on noise-free data. 
 The network parameters from the  two-stage training can actually be further refined with an end-to-end fine-tuning, which results in a slight boost. As can be seen from the training process in~Figure~\ref{end-to-end}, the  two-stages training followed by fine-tuning allows for more stable training with better results. 
In our experiments, we also found that the two-stages training is more robust and more independent from initialization. On the contrary, end-to-end training is more sensitive to initial values and is prone to training failure. 
In Figure~\ref{end-to-end} (a) and (b), we list the training records of an end-to-end model that failed to train once. 
As shown in Figure~\ref{end-to-end} (a) and (b), the end-to-end training is prone to failure due to training fluctuations, while two-stage training  results in a smooth progression. To compare the training robustness of the different schemes, we trained the lightweight end-to-end network and the two-stage network 10 times respectively. The training results are shown in Figure~\ref{end-to-end} (c). 
One can see that the end-to-end network is not stable. Its final results are not the highest value seen during the training process. There is only a 20\% success rate, while the two-stage method is very stable and the final result always reaches the highest value of each training.
This shows that, although CNNs have a powerful fitting capability that enables addressing multiple tasks in an end-to-end fashion, it is still important to consider the order of the tasks to design a reasonable pipeline.

\paragraph{Dependency on the training dataset}
In order to better compare the advantages of the network architecture regardless of the influence of training data and training strategies, we retrained  JCNN, using the same training data and training strategy as for our algorithm. Table~\ref{JCNN} shows the PSNR and SSIM  of  noise-free demosaicking and joint demosaicking and denoising with noise level  $\sigma=20$.
Among them, JCNN-O represents the original parameters of JCNN and JCNN-R stands for  the retrained version of JCNN. 
From Table \ref{JCNN}, one can see that both our proposed method and its lightweight version  outperform JCNN by a margin larger than 0.7 dB for all five test image sets under the same training data and training strategy. This means that our proposed structure is superior to  JCNN  for both tasks.

\begin{table}[t]
	\begin{center}
	\caption{Comparison of the results (PSNR/SSIM) of original JCNN (JCNN-O), retrained JCNN (JCNN-R) and the proposed method for five image sets. The best value is marked in {\textbf{bold}}, the second is marked in {\color{red}red}.}
	\label{JCNN}
	    \scriptsize
		\renewcommand\arraystretch{1.2}
		\addtolength{\tabcolsep}{-4.1pt}
		\begin{tabular}{|c|l|c|c|c|c|}
			\hline
            $\sigma$  & Dataset  & JCNN-O  & JCNN-R & Ours(-) & Ours \\ \hline
            \multirow{5}{*}{0} & Kodak & 42.09/0.9881 & 41.65/0.9874 & {\color{red}42.49}/{\color{red}0.9888} & \textbf{42.76}/\textbf{0.9893} \\
            &  McMaster & 38.95/0.9695 & 38.68/0.9677 & {\color{red}39.25}/{\color{red}0.9702} & \textbf{39.61}/\textbf{0.9725} \\
            &  WED + Flickr & 39.24/0.9807 & 39.11/0.9800 & {\color{red}39.84}/{\color{red}0.9820} & \textbf{40.22}/\textbf{0.9831} \\
            &  Urban 100 & 38.12/0.9842 & 37.97/0.9833 & {\color{red}38.88}/{\color{red}0.9852} & \textbf{39.52}/\textbf{0.9864} \\
            &  MIT moiré & \textbf{36.65}/\textbf{0.9588} & 35.19/0.9462 & 35.97/0.9516 & {\color{red}36.53}/{\color{red}0.9533} \\ \hline
            
            \multirow{5}{*}{15} & Kodak & 31.32/0.8586 & 31.37/0.8597 & {\color{red}32.12}/{\color{red}0.8780} & \textbf{32.24}/\textbf{0.8807} \\
            &  McMaster     & 31.25/0.8564 & 31.31/0.8645 & {\color{red}32.25}/{\color{red}0.8867} & \textbf{32.40}/\textbf{0.8902} \\
            &  WED + Flickr & 30.96/0.8719 & 31.09/0.8819 & {\color{red}32.07}/{\color{red}0.9022} & \textbf{32.20}/\textbf{0.9048} \\
            &  Urban 100    & 29.45/0.8912 & 29.18/0.8953 & {\color{red}30.96}/{\color{red}0.9233} & \textbf{31.21}/\textbf{0.9268} \\
            &  MIT moiré    & 28.81/0.8283 & 28.36/0.8165 & {\color{red}29.60}/{\color{red}0.8544} & \textbf{29.83}/\textbf{0.8597} \\ \hline
            
            \multirow{5}{*}{20} & Kodak & 29.91/0.8168 & 30.04/0.8228 & {\color{red}30.89}/{\color{red}0.8487} & \textbf{31.01}/\textbf{0.8518} \\
            & McMaster     & 29.79/0.8139 & 30.08/0.8338 & {\color{red}31.06}/{\color{red}0.8635} & \textbf{31.21}/\textbf{0.8670} \\
            & WED + Flickr & 29.53/0.8354 & 29.83/0.8529 & {\color{red}30.86}/{\color{red}0.8800} & \textbf{31.00}/\textbf{0.8832} \\
            & Urban 100    & 28.00/0.8552 & 27.82/0.8642 & {\color{red}29.59}/{\color{red}0.9012} & \textbf{29.86}/\textbf{0.9060} \\
            & MIT moiré    & 27.57/0.7842 & 27.27/0.7758 & {\color{red}28.47}/{\color{red}0.8213} & \textbf{28.71}/\textbf{0.8288} \\ \hline
		\end{tabular}
	\end{center}
\end{table}

\begin{table}[!t]
	\begin{center}
	\caption{Average running time of demosaicking and joint demosaicking-denoising for 500 images ($512\times512$) on a PC with Intel Core i7-9750H 2.60GHz, 16GB memory, and Nvidia GTX-1650 GPU.}
	\label{time}
	    \scriptsize
		\renewcommand\arraystretch{1.2}
		\addtolength{\tabcolsep}{-1.5pt}
		\begin{tabular}{|c|l|r|r|r|r|}
			\hline
	        &  Method       & CPU(s)  & GPU(s)  & GFLOPs    & Para(M) \\ \hline
	        \multirow{7}{*}{DM}   
			& GBTF~\cite{Pekkucuksen2010gbtf}     & 2.74    & --      & --      & --   \\ 
			& MLRI+wei~\cite{Kiku2016WMLRI}       & 1.35    & --      & --      & --   \\ 
			& ARI~\cite{monno2017adaptive}        & 25.58   & --      & --      & --   \\ 
			& CDM-CNN~\cite{tan2017color}         & 6.84    & 0.07    & 276.19  & 0.53 \\ 
			& CDM-3-Stage~\cite{cui20183-part}    & 18.61   & 0.35    & 1871.27 & 3.57 \\ 
		    & Ours(-) (DM)                        & 15.12   & 0.28    & 92.34   & 0.35 \\ 
			& Ours (DM)                           & 24.86   & 0.44    & 176.23  & 0.67 \\ \hline

			\multirow{7}{*}{JDD} 
			& ADMM~\cite{Tan2017ADMM}             & 472.27  & --      & --     & -- \\ 
			& C-RCNN~\cite{kokkinos2018deep}      & 112.77  & 2.32    & 2112.8 & 0.38\\ 
			& JCNN~\cite{gharbi2016deep}          & 10.41   & 0.22    & 53.20  & 0.56\\ 
			& LCNN-DD~\cite{Huang2018Lightweight} & 1.86    & 0.04    & 14.89  & 0.23\\ 
			& JDNDMSR~\cite{xing2021end}          & 73.15   & 1.61    & 1641.77 & 6.33\\
			& Ours(-) (DM+DN)                     & 30.13   & 0.55    & 184.68 & 0.70 \\ 
			& Ours (DM+DN)                        & 49.53   & 0.86    & 352.46 & 1.34 \\
			\hline		
		\end{tabular}
	\end{center}
\end{table}

\paragraph{Computational complexity} In order to estimate the computational complexity of these algorithms, we tested the average time consumed by all algorithms to process 500 images ($512\times512$) on a PC with Intel Core i7-9750H 2.60GHz, 16GB memory, and Nvidia GTX-1650 GPU. For the deep learning algorithms, only the actual network processing time was calculated, not including the network loading time.
The time consumed by the algorithm in demosaicking noise-free images and in demosaicking and denoising CFA images with noise level of $\sigma=10$ is shown in Table~\ref{time}. Since our network is composed of independent demosaicking and denoising stages, the time consumed can be calculated separately. In Table~\ref{time}, DM denotes the demosaicking stage of our algorithm and JDD denotes joint demosaicking and denoising.
It can be seen that the processing time of our algorithm is comparable to the other deep learning algorithms.
{It is also faster than some traditional iterative algorithms,} such as ARI~\cite{monno2017adaptive} and ADMM~\cite{Tan2017ADMM}.

\section{Conclusion}
\label{sec:Conclusion}

In this paper, we proposed a CNN for joint demosaicking and denoising. The proposed method relies on a demosaicking first then denoising approach, which is realized by applying sequentially two CNNs. 
In the first stage,  the GBTF algorithm is combined with a CNN to reconstruct a full-color image from noisy CFA image but ignoring the  image noise. In the second stage, we use another CNN to learn to remove the noise whose statistical properties were changed by the demosaicking stage. This allows to remove demosaicing noise that would otherwise be virtually impossible to remove using model-based methods.

 More importantly, we show that even when dealing with CNNs with powerful fitting capabilities a reasonable pipeline  and its training (such as the proposed two-stage training) can lead to significant performance gains with respect to more mainstream approaches based on end-to-end training. In addition, in order to improve the performance of the proposed method, we proposed an architecture based on Inception blocks as well as a lightweight version with a good speed-performance trade-off.
Experiments conducted on multiple datasets confirmed that our algorithm favourably compares to the state-of-the-art demosaicking algorithms and joint demosaicking and denoising algorithms.

\section*{Acknowledgment}
    This work was supported by National Natural Science Foundation of China (No. 12061052), Young Talents of Science and Technology in Universities of Inner Mongolia Autonomous Region (No. NJYT22090), Natural Science Foundation of Inner Mongolia Autonomous Region of China (No. 2020MS01002), Innovative Research Team in Universities of Inner Mongolia Autonomous Region (No. NMGIRT2207),  Special Funds for Graduate Innovation and Entrepreneurship of Inner Mongolia University (No.~11200-121024), Prof. Guoqing Chen's “111 project” of higher education talent training in Inner Mongolia Autonomous Region and the network information center of Inner Mongolia University.
    Work partly financed by Office of Naval research  grants N00014-17-1-2552 and N00014-20-S-B001,  DGA Defals challenge n$^\circ$ ANR-16-DEFA-0004-01, MENRT and Fondation Mathématique Jacques Hadamard.
    Y. Guo and Q. Jin are very grateful to Professor Guoqing Chen for helpful comments and suggestions. 
    The authors are also grateful to the reviewers for their valuable comments and remarks.

\bibliographystyle{elsarticle-num-names}
\bibliography{ref}

\end{document}